\documentclass[11pt]{article}

\usepackage[final]{acl}

\usepackage{times}
\usepackage{latexsym}

\usepackage[T1]{fontenc}
\usepackage[utf8]{inputenc}

\usepackage{microtype}

\usepackage{inconsolata}

\usepackage{graphicx}

\usepackage{adjustbox}

\usepackage{multirow}

\usepackage{amsmath}
\usepackage{amsfonts}

\usepackage{bbm}
\usepackage{amssymb}
\usepackage{booktabs}
\usepackage{listings}
\usepackage{xcolor}

\usepackage{listings}
\usepackage{pifont}
\usepackage{float}

\usepackage{subcaption}
\title{Collab-Overcooked: Benchmarking and Evaluating \\ Large Language Models as Collaborative Agents}

\author{
 \textbf{Haochen Sun\textsuperscript{1}}\thanks{These authors contributed equally to this work.},
 \textbf{Shuwen Zhang\textsuperscript{1}}\footnotemark[1],
 \textbf{Lujie Niu\textsuperscript{1}},
 \textbf{Lei Ren\textsuperscript{2}},
 \textbf{Hao Xu\textsuperscript{2}},
 \textbf{Hao Fu\textsuperscript{2}},\\
 \textbf{Fangkun Zhao\textsuperscript{1}},
 \textbf{Caixia Yuan\textsuperscript{1}}\thanks{Corresponding author.},
 \textbf{Xiaojie Wang\textsuperscript{1}}
\\
\\
 \textsuperscript{1}Beijing University of Posts and Telecommunications,
 \textsuperscript{2}Li Auto Inc.
\\
\texttt{\{haochen\_sun, zhangshuwen2023, lujien, yuancx, xjwang\}@bupt.edu.cn},
\\
\texttt{\{renlei3, fuhao8\}@lixiang.com}, \texttt{kingsleyhsu1@gmail.com}
}

\begin{document}
\maketitle
\begin{abstract}
Large Language Models (LLMs) based agent systems have made great strides in real-world applications beyond traditional NLP tasks. This paper proposes a new LLM-based Multi-Agent System (LLM-MAS) benchmark, Collab-Overcooked, built on the popular Overcooked-AI game with more applicable and challenging tasks in interactive environments. Collab-Overcooked extends existing benchmarks in two novel ways. First, it provides a multi-agent framework supporting diverse tasks and objectives and encourages collaboration through natural language communication. Second, it introduces a spectrum of process-oriented evaluation metrics to assess the fine-grained collaboration capabilities of different LLM agents, a dimension often overlooked in prior work. We conduct extensive experiments with 13 popular LLMs and show that, while the LLMs exhibit a strong ability in goal interpretation, there are significant shortcomings in active collaboration and continuous adaptation, which are critical for efficiently fulfilling complex tasks. Notably, we highlight the strengths and weaknesses of LLM-MAS and provide insights for improving and evaluating LLM-MAS on a unified and open-source benchmark. The environments, 30 open-ended tasks, and the evaluation package are publicly available at \url{https://github.com/YusaeMeow/Collab-Overcooked}.

% \url{https://anonymous.4open.science/r/Collab-Overcooked-E6C7}.

\end{abstract}

% --------------------------------------------------------------------------------------------------------------------
\section{Introduction}

Leveraging the remarkable zero-shot and few-shot learning abilities of Large Language Models (LLMs), LLM-based agents are demonstrating their potential in complex task decomposition and planning \cite{wang2023voyager,wang2023describe,li2024embodied}. Inspired by human collaborative behaviors in social activities, recent research demonstrates that multi-agent systems can significantly enhance task efficiency and tackle challenges surpassing single-agent capabilities \cite{li2023camel,hong2023metagpt,zhang2023building}. To effectively address complex real-world tasks, LLM-based Multi-Agent Systems (LLM-MAS) require three essential collaboration capabilities beyond goal interpretation, which include: (a) Competence boundary awareness: the ability to analyze task flows and environmental states to determine feasible actions, recognize limitations, and identify when external assistance is needed; (b) Communication: proficiency in utilizing standardized protocols for transmitting task-critical information and resource requests; and (c) Dynamic adaptation: responsiveness to collaboration requests and dynamically adjusting their action sequences accordingly and efficiently.

Given these fundamental requirements, establishing evaluation frameworks becomes crucial for assessing LLM-MAS collaboration effectiveness. Researchers have developed specialized benchmarks to quantify collaborative agents in specific environments. Representative platforms like \citet{agashe2023evaluating}, RocoBench \cite{mandi2024roco}, and LLMARENA \cite{chen2024llmarena} create virtual scenarios requiring collaborative problem-solving through intricate workflows. These frameworks are complemented by novel metrics, such as Collaboration Score (CoS) \cite{gong2023mindagent}, which evaluates end-to-end collaboration capability.

\begin{table*}
  \centering
  \adjustbox{max width=\textwidth}{

\begin{tabular}{lccccc}
\hline
\textbf{Virtual Environment} & \textbf{\begin{tabular}[c]{@{}c@{}}Various Task\\ Complexities\end{tabular}} & \textbf{Scalability} & \textbf{\begin{tabular}[c]{@{}c@{}}Collaboration\\ Definition\end{tabular}} & \textbf{\begin{tabular}[c]{@{}c@{}}Forced\\ Collaboration\end{tabular}} & \textbf{\begin{tabular}[c]{@{}c@{}}Collaboration\\ Evaluation\end{tabular}} \\ \hline
RocoBench \cite{mandi2024roco} & NA/6 & \ding{55} & NA & Partial & E2E \\
VillagerBench \cite{dong2024villageragent} & 3/9 & \ding{55} & E2E & \ding{55} & E2E \\
LLMARENA \cite{chen2024llmarena} & NA/7 & \ding{55} & PO & \ding{55} & E2E \\
CivRealm \cite{qi2024civrealm} & NA/100k & \checkmark & NA & \ding{55} & E2E \\
BattleAgentBench \cite{wang2024battleagentbench} & 3/3 & \ding{55} & E2E & \ding{55} & E2E \\
TDW-MAT \cite{zhang2023building} & NA/2 & \ding{55} & E2E & \ding{55} & E2E \\
CuisineWorld \cite{gong2023mindagent} & 13/39 & \checkmark & E2E & \ding{55} & E2E \\
\textbf{Collab-Overcooked} (Ours) & 6/30 & \checkmark & PO & \checkmark & E2E\&PO \\ \hline
\end{tabular}
}
\caption{\label{tab:related}
    Statistics of existing benchmarks for evaluating LLM-MAS collaboration. If no data is available, it is marked as ``NA''. Statistics in ``Various Task Complexities'' are presented in the format ``Number of Levels / Total Number of Tasks''. ``E2E'' refers to end-to-end, while ``PO'' refers to process-oriented.
  }

\end{table*}

Despite recent progress in evaluating LLM-MAS collaboration capability, existing approaches exhibit three critical limitations. First, they prioritize task completion efficiency without imposing strict collaboration requirements, allowing individual agents to accomplish tasks that are nominally ``collaborative'' independently. This design flaw introduces assessment biases by obscuring the role of collaboration in performance gains, which contrasts with real-world applications where collaboration is often essential for task success. Second, existing benchmarks conflate collaboration capability with end-to-end metrics, such as task completion rates, which are frequently used as proxies for collaboration effectiveness in platforms like CuisineWorld \cite{gong2023mindagent} and VillagerBench \cite{dong2024villageragent}. However, this approach overlooks two critical issues: divergent definitions of ``success'' across environments undermine comparability, and the absence of process-oriented metrics obscures actionable insights for optimizing collaborative strategies. Third, the lack of a fine-grained evaluation prevents a comprehensive, multi-perspective analysis of LLM agents' capabilities, making it difficult to interpret their strengths and limitations effectively, thus falling short of insightful research suggestions.

To address the limitations of existing LLM-MAS benchmarks, we propose the Collab-Overcooked Benchmark, designed to provide a fine-grained analysis of collaborative interactions. Unlike prior benchmarks that focus primarily on task completion, our benchmark evaluates the capability of initiating and responding to collaboration during the collaboration process. Specifically, the Collab-Overcooked extends Overcooked-AI \cite{carroll2019utility} to a chef-and-assistant collaborating environment and introduces 30 sequential process-specific tasks across 6 complexity levels. Each agent operates in an isolated environment with distinct action spaces, so task completion depends on effective communication and resource exchange, therefore collaboration is strictly required. Furthermore, we propose the Trajectory Efficiency Score (TES) and Incremental Trajectory Efficiency Score (ITES) to assess the collaboration capabilities from both coarse and fine perspectives. Through comprehensive experiments on 13 LLMs of varying sizes, including both open-source and closed-source LLMs, we reveal significant performance gaps in collaboration capabilities across different LLMs. We identify attention misalignment as a key factor affecting collaboration performance. Our results show that, in collaborative tasks, correcting attention alone can improve outcomes, revealing core limitations of current LLM-MAS and pointing to future directions such as collaborative memory and attention-guided fine-tuning.

To summarize, our contributions are as follows:
\begin{itemize} 
\item[$\bullet$] We develop and open-source a lightweight and extensible LLM-MAS benchmark, Collab-Overcooked, which features 30 tasks across 6 complexity levels that encourage collaboration, thus facilitating the evaluation of MAS collaboration in a unified environment with diverse, complex tasks.

\item[$\bullet$] We define collaboration capability in LLM-MAS as comprising both initiating collaboration and responding to collaboration. We introduce 3 trajectory efficiency-related metrics to evaluate collaboration capabilities from both coarse and fine-grained perspectives.

\item[$\bullet$] We conduct a comprehensive evaluation of a wide range of popular LLM agents, revealing collaboration and adaptation bottlenecks under varying task complexities, and identifying key limitations of LLM-MAS through analysis of attention distribution.
\end{itemize}

% --------------------------------------------------------------------------------------------------------------------

\section{Related Work}

\paragraph{LLM-Based Multi-Agent System}

LLM-MAS enables agents to collaboratively engage in planning, discussing,  and decision-making. Collaboration is a pivotal capability in task-oriented LLM-MAS, as it not only enhances task completion efficiency \cite{zhang2024towards,tao2024github} but also enables the pursuit of complex goals beyond the reach of a single agent \cite{park2023generative,hong2023metagpt}. Recent methods for improving collaboration can be broadly categorized into (a) Structural optimization (e.g., DyLAN’s \cite{liu2023dynamic} dynamic framework), (b) Role specialization (e.g., AutoGen’s \cite{wu2023autogen} personas and AgentVerse's \cite{chen2023agentverse} role assignments), and (c) Communication paradigm (e.g., MetaGPT's \cite{hong2023metagpt} message pool). Despite these advancements, the inherent complexity and diversity of multi-agent tasks make it difficult to compare methods directly, driving the emergence of standardized benchmarks that enable quantitative evaluations under unified conditions.

\paragraph{LLM-MAS Benchmark and Evaluation}

Benchmark testing in virtual environments is the primary method for evaluating multi-agent collaboration capability. As shown in Table \ref{tab:related}, existing studies establish diverse tasks and commonly use End-to-End (E2E) metrics to assess LLM-MAS collaboration capability, with some benchmarks offering environmental scalability. However, several limitations persist. A key issue is the lack of a formal collaboration definition in most benchmarks, leading to ambiguous assessments and inconsistent comparisons across different benchmarks. Furthermore, the absence of enforced collaboration mechanisms allows agents to achieve objectives independently (e.g., in CuisineWorld, where many tasks can be completed by a single agent), undermining the true assessment of collaboration. Finally, the predominant focus on outcome-based metrics such as E2E performance overlooks the critical role of process-driven dynamics. Approaches like \cite{song2024trial}, LTC \cite{wang2023ltc}, and EvoMAC \cite{hu2024evomac} suggest refining LLMs through process behaviors to enhance adaptation and collaboration, indicating that incorporating process-oriented metrics could offer more comprehensive insights.

\begin{figure*}[t]
  \includegraphics[width=1\linewidth]{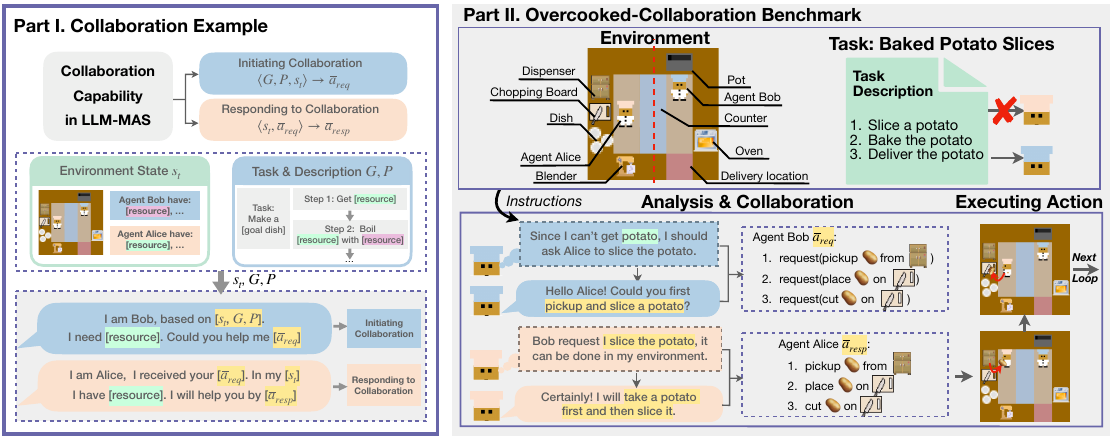}
  \caption {Part I presents the collaboration process, which is divided into initiating collaboration and responding to collaboration. Part II outlines the design of the Collab-Overcooked Benchmark, emphasizing its characteristics of resource isolation and asymmetric task knowledge, and provides an example of agents' collaboration.
}
  \label{fig:benchmark}
\end{figure*}

% --------------------------------------------------------------------------------------------------------------------

\section{Task-Oriented Collaboration}

\subsection{Collaboration Capability}

A task in LLM-MAS can be formulated as a 4-tuple:  $ T = (G, E, \mathcal{P}, \mathcal{R})$, where $ G $ is a natural language description of the task goal, such as ``make a dish of tomato soup''; $ E $ is a description of the environment, which can be either the layout of a simulated scenario or the visual input of real-world surroundings; $ \mathcal{P} $ is optional natural language guidance, providing recipes, helpful hints, or task constraints; and $ \mathcal{R} $ is a Referential Action Trajectory (RAT) that leads to the successful completion of the task and is used to assess the agents' performance. It is worth noting that there are often multiple RATs for a task, especially in dynamic environments.

Collaboration often involves agents relying on each other to solve tasks. As shown in Figure \ref{fig:benchmark} Part I, we define collaboration capability as comprising two essential components: the capability to initiate collaboration, where agents, upon realizing that their boundary prevents them from completing the task according to $ G $ and $ \mathcal{P} $ at environmental state $ s_t \in E $ at time $ t $, generate a request for collaborative actions $ \overline{a}_{req} $ to solicit assistance from other agents; and the capability to respond to collaboration, where agents, upon receiving $ \overline{a}_{req} $ from another agent, adjust their action sequence based on $ s_t $ and generate collaborative actions $ \overline{a}_{resp} $.

\subsection{TES and ITES}
\subsubsection{TES}
 
Trajectory Efficiency Score (TES) is designed to compare the difference between two trajectories and is defined as:

\begin{equation}
  \label{eq:sea}
\text{TES}(\overline{h}_k) = \max_j\left\{\frac{(1+\beta^2)D_{\text{max}}^j(\overline{h}_k, \overline{g}_k^j)}{m_k + \beta^2 n_k}\right\}
\end{equation}

\noindent where $\overline{h}_k= \{a_k^1, a_k^2, \ldots, a_k^T\}$ is the historical action sequence up to timestep $T$ of agent $k$, $\overline{g}_k^j=\{g_i\}_{i=1}^{m_k} \in \mathcal{R}$ is $j$-th RAT of agent $k$, $\beta$ is the hyperparameter balancing the weight of task progress and redundancy, and $D_{\text{max}}^j(\overline{h}_k, \overline{g}_k^j)$ computes the length of the longest order-preserving subsequence in $\overline{h}_k$ that matches $\overline{g}_k^j$:  

\begin{multline}
  \label{eq:dmax}
     D_{max}^j = \max_d \{ d \mid \forall \, 1 \leq i_1 < \dots < i_d \leq n_k, \, \\
     \text{s.t.} \ a_{i_1} = g_1, a_{i_2} = g_2, \dots, a_{i_d} = g_d \}
\end{multline}

Unlike other sequence alignment scores (such as ROUGE-L \cite{lin2004rouge}), TES takes into account sequence order and a redundancy penalty simultaneously, therefore suitable for assessing a planned action sequence (detailed in Appendix~\ref{ap-sea}).

\subsubsection{ITES}

Incremental Trajectory Efficiency Score (ITES) introduces an incremental assessment to quantify the task-progress contribution of an individual collaborative action. The ITES is computed as:

\begin{equation}
  \label{eq:isea1}
\text{ITES}(\overline{a}, \overline{h}_k) = \text{TES}(\overline{h}_k \cup \overline{a}) - \text{TES}(\overline{h}_k)
\end{equation}

\noindent where $\overline{h}_k$ denotes the historical action sequence of agent $k$, and $\overline{a}$ represents the newly executed actions, either a collaboration request ($\overline{a}_{req}$) or response ($\overline{a}_{resp}$).

This differential formulation measures the marginal utility of action $\overline{a}$ by evaluating its impact on trajectory alignment with the RATs. It can be established that: $\text{ITES}(\overline{a}, \overline{h}_k) > 0$ indicates $\overline{a}$ advances task progress, $\text{ITES}(\overline{a}, \overline{h}_k) \leq 0$ suggests $\overline{a}$ fails to advance task progress (i.e., $\overline{a}$ is redundant / premature action or incorrect response).

%  (e.g., valid resource requests or correct responses)

\subsection{Evaluation Metrics}

\paragraph{Progress Completeness (PC)} 

Built on the TES, which quantifies a piece of trajectory, PC measures the task progress of all involved agents while penalizing redundancy as a whole, and is defined as:

\begin{equation}
    PC = \frac{1}{K}\sum_{k=1}^K\text{TES}(\overline{h}_k) 
\end{equation}

\noindent where $K$ is the number of agents, $\overline{h}_k = \bigcup_{t=0}^{T_{max}}a_k^t$ denotes the historical action sequence of agent $k$ at time $T_{max}$, which occurs upon task completion or when the maximum time limit is reached. The PC offers a finer-grained assessment of task completion efficiency compared to boolean success rate.

\paragraph{Initiating Capability (IC)}

IC evaluates the correctness of the LLM agent’s collaboration initiation. IC is defined as:  

\begin{equation}
    IC= \frac{1}{N} \sum_{i=1}^{N} \mathbb{I}\left( \text{ITES}(\overline{a}_{req}^{(i)}, \overline{h}_j) > 0 \right)
\end{equation}

\noindent where $N$ is the number of required collaborations, $\mathbb{I}()$ is the indicator function. $\mathbb{I}\left( \text{ITES}(\overline{a}_{req}^{(i)}, \overline{h}_j) > 0 \right)$ determines whether the $i$-th initiating collaboration request $\overline{a}_{req}^{(i)}$ advances the task progress, thereby indicating whether the initiation is correct. 

\paragraph{Responding Capability (RC)}

Similarly, RC assesses the correctness of the LLM agent's response to a collaboration request:

\begin{equation}
    RC= \frac{1}{N} \sum_{i=1}^{N} \mathbb{I}\left( \text{ITES}(\overline{a}_{resp}^{(i)}, \overline{h}_j) > 0 \right).
\end{equation}

% --------------------------------------------------------------------------------------------------------------------

\section{Benchmark}

\subsection{Collab-Overcooked Benchmark}

The proposed Collab-Overcooked benchmark builds upon the open-source Overcooked-AI \cite{carroll2019utility} and ProAgent \cite{zhang2024proagent}, introducing two key upgrades: (1) The environment is divided into two parts, featuring resource isolation and asymmetric task knowledge for Agent Bob and Agent Alice, respectively. This contrasts with Overcooked-AI, where agents mostly operate in a shared environment with identical items; (2) The benchmark encourages collaboration through natural language interactions, with some cases enforcing collaboration as a requirement for task success. Additionally, Collab-Overcooked provides APIs to configure new tasks and environmental settings, enabling the enhancement of LLM-MAS through scenario adaptation.

\subsubsection{Environment}

Our simulation environment is a grid-based kitchen simulation designed as a comprehensive testbed for analyzing collaboration behaviors in LLM-MAS. The environment comprises agents and configurable interactive elements. The interactive elements are dispensers, utensils, counters, and delivery location. Agents can freely retrieve raw materials from dispensers, place them into utensils for processing, and finally transfer the processed materials to other agents via counters or submit the required order through the delivery location. Notably, utensils process materials according to customizable synthesis tables, with each utensil having its own distinct synthesis table. Agents can interact with these elements through predefined action primitives formatted as ``func(args)''. For example, ``pickup(apple, ingredient\_dispenser)'' clarifies action type, target material, and interactive element. Details are provided in Appendix~\ref{ap-environment}.

The environment executes agents' actions sequentially and broadcasts the global state at each timestep, encompassing agents' positions and the status of interactive elements. We developed a comprehensive rule-based action validator that identifies invalid actions, including environment-action mismatches and incorrect parameters. Upon rule violations, the validator issues error messages, prompting the agent to identify the error and regenerate the action accordingly.

\subsubsection{Tasks Construction}

Sequential process-specific tasks are common in real-world scenarios \cite{wang2023describe,zhang2023building,song2024trial}, where interdependent actions must be completed in a specific order to achieve a goal. We curate 30 such tasks stratified into 6 complexity levels, requiring two agents to complete collaboratively. The task complexity level is determined by the minimum number of collaborative actions, increasing linearly with difficulty. To reduce LLM bias toward specific ingredients, tasks at the same level share workflows but differ in ingredients. Each task has a time constraint, set as the optimal completion time scaled by a time limit factor $\gamma$.

Each task is accompanied by a natural language structured process description and RATs for evaluation. As the tasks are process-specific with clear success criteria, their RATs are fully definable and easily traversable, making them suitable for evaluation. We manually annotated RATs for all 30 tasks. Detailed task list, task descriptions, and RAT examples are provided in the Appendix~\ref{ap-task-construction}.

\subsubsection{Collaboration Designs}

Collab-Overcooked benchmark imposes strict collaboration among agents. For this, we have two special designs: (a) Resource Isolation: agents operate in resource-isolated sub-environments, necessitating resource exchange via a shared ``counter''. This enforces collaborative dependency. (b) Asymmetric Task Knowledge: Only one agent knows how to complete the task. Agents must communicate to synchronize task information. This setup makes collaboration indispensable, offering a stronger test of collaborative capabilities and better exposing potential deficiencies in collaboration.

While our current setup uses two agents to clearly expose and evaluate collaboration initiation and response, scaling to multiple agents primarily introduces complexity in collaboration rather than fundamentally altering these core collaborative capabilities. Thus, the two-agent design is optimal for isolating and analyzing LLM-specific collaborative behaviors in depth.

\subsection{Baseline}
\label{sec-baseline}
To evaluate LLM-MAS performance across different LLMs on our benchmark, we introduce an in-context learning baseline. The baseline incorporates both memory and reflection mechanisms, allowing agents to communicate and collaborate freely in natural language while handling errors. Notably, our baseline architecture is aligned with common agent architectures \cite{dong2024villageragent,zhao2024expel,zhu2025multiagentbench}. Figure \ref{fig:benchmark} Part II illustrates an example of how agents advance task progress through collaborative communication in our benchmark. Additionally, we provide prompts in detail, which include the game rules, communication formats, and action space definitions, as well as error correction and reflection procedures. Detailed information can be found in Appendix \ref{ap-baseline} and Figure \ref{fig:appendix-baseline}.

% 

% --------------------------------------------------------------------------------------------------------------------
\begin{table*}
  \centering
  \adjustbox{max width=\textwidth}{

\begin{tabular}{clcccccccccccc}
\hline
\multicolumn{1}{l}{} &  & \multicolumn{2}{c}{Level 1} & \multicolumn{2}{c}{Level 2} & \multicolumn{2}{c}{Level 3} & \multicolumn{2}{c}{Level 4} & \multicolumn{2}{c}{Level 5} & \multicolumn{2}{c}{Level 6} \\
\multicolumn{1}{l}{} &  & SR & PC & SR & PC & SR & PC & SR & PC & SR & PC & SR & PC \\ \hline
\multirow{5}{*}{\begin{tabular}[c]{@{}c@{}}Closed \\ Source\end{tabular}} & GPT-4o & 94.00 & 85.92 & 86.00 & 84.96 & 68.00 & 76.61 & 34.00 & 44.42 & 2.00 & 29.13 & 4.00 & 22.45 \\
& Claude Sonnet 4 & \textbf{100.00} & \textbf{96.00} & \textbf{100.00} & \textbf{98.67} & \textbf{96.00} & \textbf{95.82} & \textbf{92.00} & \textbf{94.48} & \textbf{74.00} & \textbf{78.15} & \textbf{58.00} & 60.69 \\
& o4-mini & 92.00 & 90.93 & \textbf{100.00} & 89.60 & \textbf{96.00} & 86.15 & 86.00 & 88.39 & 62.00 & 68.59 & 54.00 & \textbf{60.79} \\
 & o1-mini & 70.00 & 74.18 & 2.00 & 36.36 & 0.00 & 33.60 & 0.00 & 24.80 & 0.00 & 20.28 & 0.00 & 13.07 \\
 & GPT-3.5-turbo & 42.00 & 68.20 & 8.00 & 43.42 & 0.00 & 36.44 & 0.00 & 24.74 & 0.00 & 15.21 & 0.00 & 12.03 \\ \hline
\multirow{7}{*}{\begin{tabular}[c]{@{}c@{}}Open\\ Source\end{tabular}} & DeepSeek-R1 & \multicolumn{1}{l}{\textbf{100.00}} & \multicolumn{1}{l}{\textbf{96.53}} & \multicolumn{1}{l}{\textbf{100.00}} & \multicolumn{1}{l}{\textbf{94.40}} & \multicolumn{1}{l}{\textbf{98.00}} & \multicolumn{1}{l}{\textbf{91.10}} & \multicolumn{1}{l}{\textbf{82.00}} & \multicolumn{1}{l}{\textbf{82.75}} & \multicolumn{1}{l}{\textbf{44.00}} & \multicolumn{1}{l}{\textbf{49.79}} & \multicolumn{1}{l}{\textbf{30.00}} & \multicolumn{1}{l}{\textbf{48.33}} \\
 & DeepSeek-V3 & 88.00 & 77.74 & 76.00 & 71.90 & 56.00 & 66.61 & 22.00 & 50.01 & 4.00 & 30.41 & 6.00 & 33.44 \\
 & Qwen2.5-72B-Instruct & 78.00 & 76.84 & 64.00 & 68.00 & 14.00 & 46.88 & 8.00 & 30.80 & 0.00 & 22.67 & 0.00 & 18.45 \\
 & Qwen2.5-32B-Instruct & 64.00 & 73.36 & 44.00 & 62.02 & 14.00 & 40.08 & 4.00 & 33.78 & 2.00 & 22.16 & 0.00 & 18.93 \\
 & Qwen2.5-14B-Instruct & 32.00 & 50.36 & 4.00 & 26.66 & 0.00 & 24.41 & 0.00 & 19.00 & 0.00 & 14.14 & 0.00 & 14.27 \\
 & Qwen2.5-7B-Instruct & 8.00 & 44.79 & 0.00 & 13.00 & 0.00 & 9.29 & 0.00 & 8.35 & 0.00 & 5.57 & 0.00 & 4.51 \\
 & Llama-3.1-70B-Instruct & 70.00 & 75.42 & 42.00 & 63.15 & 22.00 & 54.58 & 6.00 & 45.04 & 0.00 & 29.77 & 0.00 & 17.69 \\
 & Llama-3.1-8B-Instruct & 4.00 & 33.03 & 0.00 & 15.49 & 0.00 & 12.33 & 0.00 & 11.24 & 0.00 & 9.05 & 0.00 & 7.45 \\ \hline
\end{tabular}
}
\caption{\label{tab:overall}
    Performance of 13 representative LLMs with parameter sizes ranging from 7B to 671B+ across 6 task complexity levels, evaluated using Success Rate (SR) and Progress Completeness (PC) as metrics.
  }

\end{table*}

\section{Experiment and Analysis}

\begin{figure}[t]
  \includegraphics[width=\columnwidth]{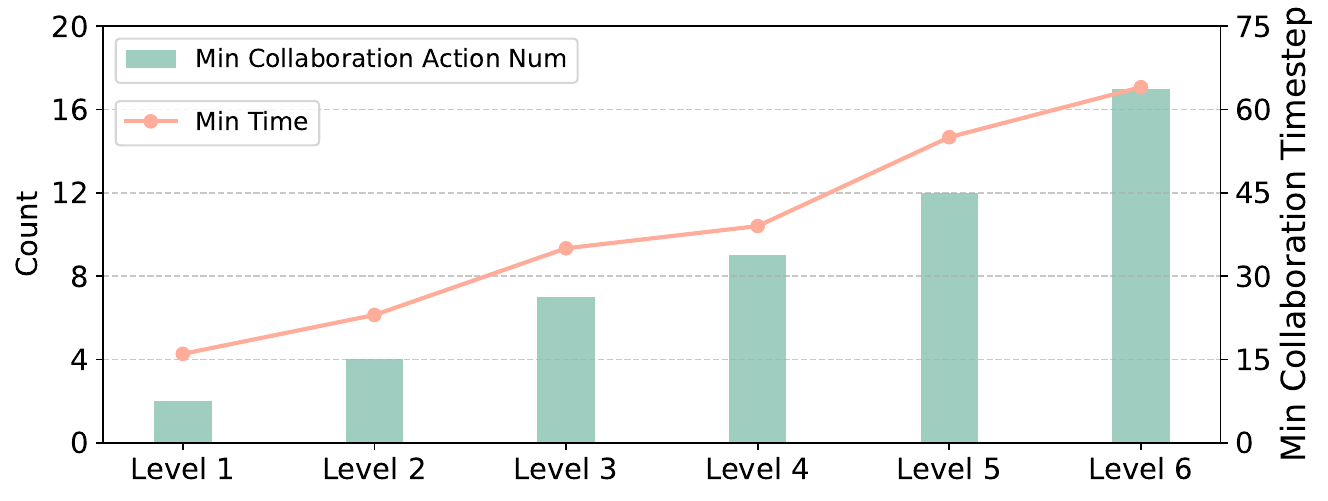}
  \caption{The statistics for tasks of varying complexity levels. ``Min Collaborative Action Num'' denotes the minimum number of collaborative actions performed by the responding agent. ``Min Time'' represents the shortest timesteps to complete a task at a given level.}
  \label{fig:benchmark-stat}
\end{figure}

\subsection{Benchmark Overview}

Figure \ref{fig:benchmark-stat} presents key statistics of our benchmark, summarizing the minimum completion timesteps and collaborative actions across 6 complexity levels, which show monotonically increasing trends with task complexity. Two agents perform 8 and 6 actions, respectively. The environment layout indicates asymmetric interactivity, with two agents accessing 4 and 5 interactive elements, respectively, while sharing observation. Additional statistics are provided in Appendix \ref{ap-environment}.

\subsection{Experiment Setting}

We leverage 13 representative LLMs with parameter sizes ranging from 7B to over 671B+ as the foundation models for LLM-MAS. The open-source models include DeepSeek-R1 \cite{guo2025deepseek}, DeepSeek-V3 \cite{liu2024deepseek}, different parameter versions of Qwen2.5 (7B, 14B, 32B, 72B) \cite{yang2024qwen2} and Llama-3.1 (8B, 70B) \cite{dubey2024llama}, all with instruction-tuned configurations. The closed-source models include: GPT-4o-1120 \cite{gpt-4o}, Claude Sonnet 4 \cite{claude4}, o4-mini \cite{o4mini}, o1-mini \cite{o1mini}, and GPT-3.5-turbo-0125 \cite{ouyang2022training}. For the open-source models except for DeepSeek-R1 and V3, inference is performed using vLLM \cite{kwon2023vllm} with temperature of 0.7 and top-p of 1. For each task, the task time limit factor is set to $\gamma=1.5$\footnote{Experiments for different $\gamma$ are in Appendix \ref{ap-gamma}.}, and each task is evaluated through 10 repetitions. The hyperparameter $\beta$ in TES is 0.95. To ensure a consistent and fair evaluation, we employ the identical agent architecture for all models, presented in Section \ref{sec-baseline}.

\begin{figure*}[t]
  \includegraphics[width=1\linewidth]{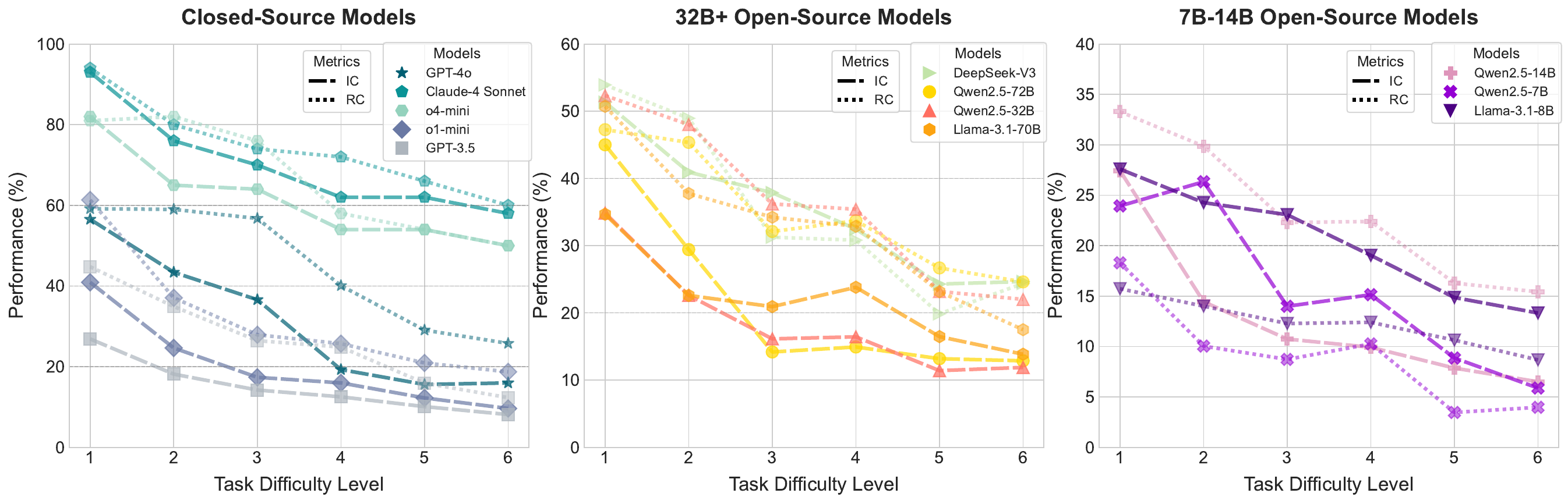}
  \caption {The performance of 13 representative LLMs, with parameter sizes ranging from 7B to 671B+, was evaluated across 6 task levels using the IC, and RC.}
  \label{fig:collab_metric}
\end{figure*}

\begin{figure}[t]
  \includegraphics[width=\columnwidth]{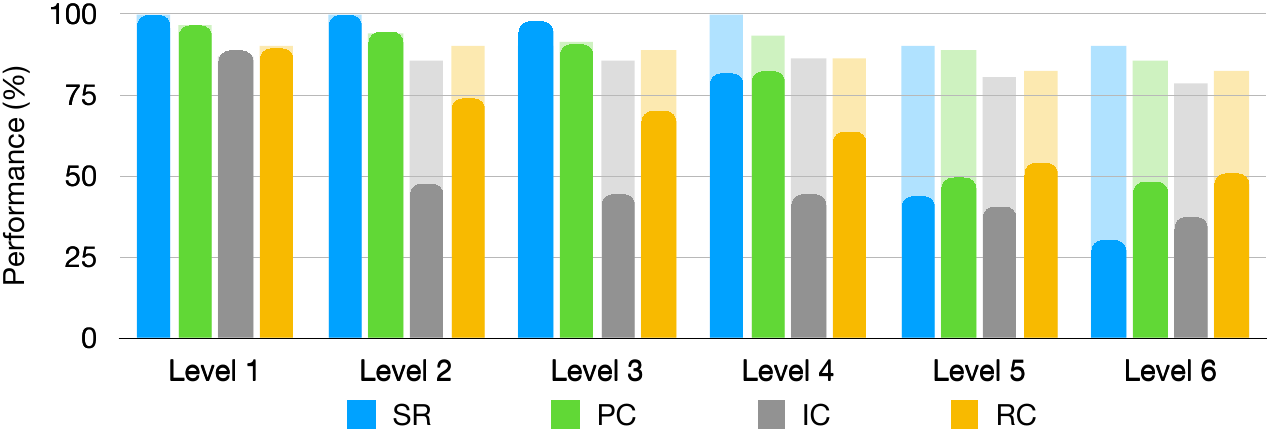}
  \caption{Comparison of human performance (shown as lighter, semi-transparent bars, under a 10-second per-timestep thinking and communication constraint) and DeepSeek-R1 performance (shown as darker, solid bars) across six task complexity levels in our benchmark.}
  \label{fig-human-result}
\end{figure}

\subsection{Results and Analysis}
\label{result}

\subsubsection{Task Completion Efficiency}

Table~\ref{tab:overall} presents the Success Rate (SR) and PC scores of 13 LLMs across six levels. Claude Sonnet 4 demonstrates the strongest overall performance, consistently outperforming other models on higher-complexity tasks. Among the open-source models, DeepSeek-R1 excels, especially on tasks of low to medium complexity. However, its token usage is 18.6 times that of GPT-4o, indicating a significant computational trade-off. From these results, we derive three key insights: (1) Smaller LLMs (8B parameters or fewer) struggle with simple tasks, whereas increasing model size significantly enhances performance. This suggests the existence of a clear emergent scaling threshold for low-level tasks. (2) Scaling up LLMs effectively improves task completion efficiency for lower-level tasks but fails to enhance performance on high-complexity tasks. This suggests that current performance gains primarily stem from pattern memorization rather than cognitive reasoning. (3) Beyond Level 4, model performance diverges sharply. Most models collapse, while top-tier agents like Claude Sonnet 4 and o4-mini only postpone this failure to higher complexity levels. This trend indicates that even the most capable models eventually falter under increasing collaborative demands, highlighting our benchmark's challenging nature and the persistent difficulty of long-horizon reasoning.

\subsubsection{Process-Oriented Evaluation}

Figure~\ref{fig:collab_metric} presents the process-oriented evaluation of LLM-MAS, from which we derive three key insights. First, most models (14B+) exhibit higher RC than IC, indicating that LLMs are better at responding to collaboration than initiating collaboration. This is a result of their strong instruction-following capabilities, which make initiating collaboration the primary bottleneck for most LLMs. Second, the collaboration capability of all LLMs declines with increasing task complexity. Moreover, the decline rate is similar across all models, indicating that their ability to maintain collaboration performance is similar. Despite the scale-up of the models, there is no corresponding improvement in their ability to sustain collaboration capability. Third, the reasoning model outperforms others on simpler tasks. While its performance drops with complexity and it consumes more tokens, its consistent gains show the potential of the CoT-training paradigm for improving collaboration capabilities.

\begin{figure*}[t]
  \includegraphics[width=1\linewidth]{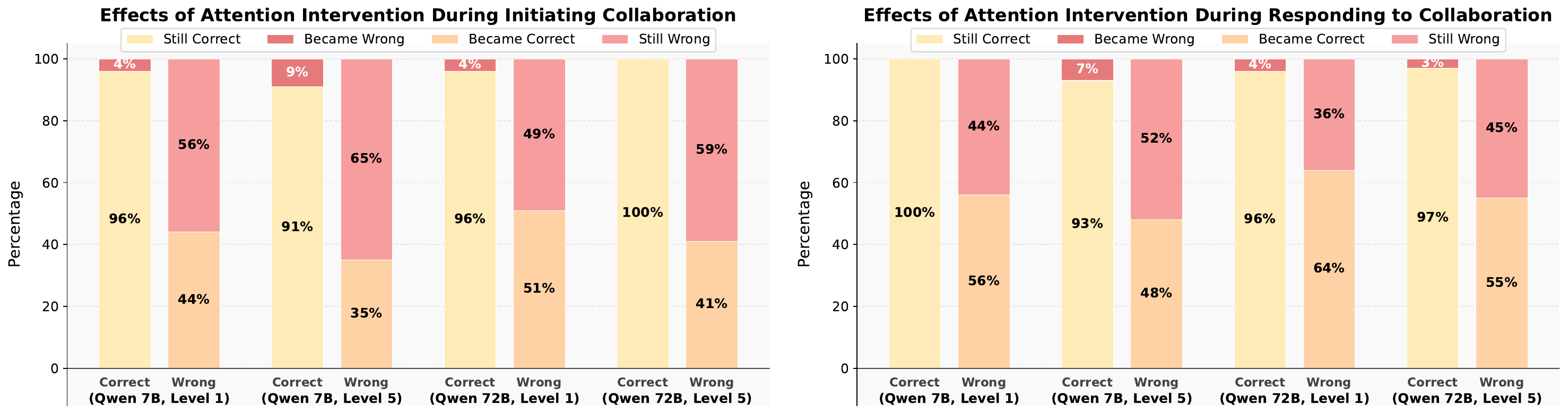}
  \caption {Results for Qwen-2.5 7B and 72B on Level 1 and Level 5 tasks. The left panel shows initiating collaboration, and the right shows responding. ``Correct'' and ``Wrong'' indicate the model's original output before intervention. Results show that manually guiding the model's attention to align with successful patterns can significantly improve performance on previously incorrect cases while maintaining stability on correct ones.
}
  \label{fig:attention}
\end{figure*}

\subsubsection{Human Performance Evaluation}

To establish a robust performance ceiling, we conducted experiments with 10 human participants performing tasks spanning all six levels. As shown in Figure \ref{appendix:human-alice}, \ref{appendix:human-bob}, we designed a human-computer interaction interface to enable participants to simulate agent behaviors within the environment. To ensure a fair comparison with LLMs in time-sensitive scenarios, we imposed time constraints on both communication and decision-making during each timestep for participants. We further evaluated human performance under various time limits, and detailed descriptions of the experimental design and rules are provided in Appendix \ref{ap-human}.

As illustrated in Figure~\ref{fig-human-result}, human participants consistently achieved high and stable performance across all levels of task complexity, even under time constraints. In contrast, DeepSeek-R1, the strongest open-source model evaluated, exhibited a marked decline in performance as task complexity increased. These results highlight two key limitations of current LLM-MAS: a lack of performance consistency under increasing complexity, and the insufficiency of model parameter scaling alone to overcome this gap. This advantage in human performance stems from the participants’ ability to form high-level task abstractions and procedural understanding during interaction, allowing them to flexibly adapt to novel situations and maintain stable outcomes. In comparison, current LLM-MAS rely on shallow memory mechanisms that log past trajectories without abstracting them into reusable strategies. Consequently, they fail to generalize from simpler tasks to more complex ones, leading to cumulative errors and performance degradation as complexity rises.

\subsubsection{Analysis of Collaboration Failures}
\label{fa}

Collaboration challenges are likely to arise across diverse multi-agent contexts. Recent work has begun to systematically categorize the failure modes of LLM-MAS. For example, \citet{cemri2025multi} proposed MAST, a comprehensive taxonomy of MAS failures. While such taxonomies provide a valuable framework for understanding what failures occur, our benchmark's design, with its strict resource isolation and asymmetric task knowledge, allows a deeper, quantitative analysis into why and how these failures manifest dynamically.

\paragraph{Collaboration Capabilities Degradation}

To better understand collaboration capability degradation, we conducted a series of controlled experiments detailed in Appendix~\ref{ap-fa}. We first identify a consistent performance decay across sequential collaborative steps. As tasks progress, agents become increasingly prone to specific failure modes such as premature or repetitive initiation, establishing the capability of initiating collaboration as the primary bottleneck. This trend of performance degradation in long-horizon tasks is consistent with findings from \citet{li2024measuring} and \citet{li2024autobencher}. Crucially, our analysis reveals this degradation persists even when we mitigate planning ambiguity by providing recipes with explicit step-to-action mappings. This experimental control allowed us to isolate the root cause to a strong positional dependence: identical collaborative actions that fail in later stages are executed with significantly higher success rates when their position is simply moved to the beginning of the workflow. This finding strongly suggests that the issue is not a failure in high-level reasoning, but stems from more fundamental limitations: inherent pretraining biases that favor sequence initiation over continuation, and an architectural inability to maintain coherent context over extended interactions, which are critical for sustained collaboration.

\paragraph{Attention Bias}

By segmenting input prompts into 5 or 6 distinct parts and analyzing attention weight distributions (see Figure~\ref{appendix-prompt-part}), we identified distinct attention patterns differentiating successful and failed collaborations, highlighting critical biases. During initiation, increased attention to collaboration rules correlates with success, whereas excessive focus on recipe information predicts failure. This suggests a fundamental attention bias where LLM-MAS agents overemphasize task execution details while undervaluing essential collaboration-specific information, causing errors in determining the appropriate collaboration approach at a given state (see Appendix~\ref{appendix-corr-supp}). In the responding phase, successful outcomes feature heightened attention to environmental observations and collaboration rules. In contrast, excessive deference to partner instructions without integrating environmental observation and collaboration rules causes failed responding. These attention biases directly contribute to redundant actions and degraded performance metrics (PC, IC, and RC), with their effects becoming more pronounced under increased task complexity due to error propagation.

\paragraph{Attention Intervention}
To establish the causal relationship between attention distribution and collaboration outcomes, we conducted attention intervention experiments by manually adjusting the attention allocation to align with patterns observed in successful cases. Using the same random seeds and model parameters, we then regenerated the outputs. As shown in Figure~\ref{fig:attention}, we observed performance improvements of 35\% to 64\% in previously failed instances, while originally successful outputs remained largely unaffected. These results confirm that attention bias is a key causal factor in collaboration failure, likely rooted in pretraining on single-agent execution tasks rather than on collaborative scenarios requiring joint decision-making.

To the best of our knowledge, we are the first to reveal and analyze attention-driven failure modes in information and resource isolation environments, highlighting persistent biases toward task execution that are less evident in existing LLM-MAS collaboration benchmarks.

\subsection{Future Challenges}
\paragraph{Collaborative Memory and Experience Abstraction}
Future work should develop specialized memory mechanisms for multi-agent collaboration that go beyond single-agent approaches. LLM-MAS requires systems that can retain and generalize collaborative patterns across diverse contexts and complexity levels, enabling agents to progressively develop more sophisticated collaboration capabilities through accumulated experience.

\paragraph{Attention-Guided Fine-tuning}
Our attention intervention experiments demonstrate that targeted attention modification alone can dramatically improve collaborative outcomes. Future approaches should incorporate mechanisms that guide models to attend to critical collaboration-relevant information through fine-tuning regimens or soft attention constraints. These techniques could help overcome the inherent single-agent execution biases currently limiting LLM collaborative performance.

% --------------------------------------------------------------------------------------------------------------------

\section{Conclusion}

We introduce the Collab-Overcooked Benchmark, a framework evaluating LLM-MAS collaboration from end-to-end and process-oriented perspectives. Experiments across 13 LLMs reveal significant performance gaps, with attention misalignment to collaboration-relevant instructions emerging as a key bottleneck. These findings underscore the difficulty of achieving high performance in collaborative tasks under training-free, zero-shot settings, highlighting the need to improve attention mechanisms for better adaptability and collaboration.

% --------------------------------------------------------------------------------------------------------------------

% \clearpage

\section*{Limitations}

The Collab-Overcooked Benchmark is introduced in our paper and we explore methods for evaluating the collaboration capabilities of LLM-MAS using both end-to-end and process-oriented approaches. However, there are three limitations to our work. First, all of our tasks are sequential and process-specific. While we assume that RATs can be exhaustively enumerated, making it possible to use exhaustive RATs as labeled data for evaluating the collaboration capabilities of LLM-MAS. However, in environments with highly complex state and action spaces, RATs are difficult to exhaustively enumerate. In such cases, only representative RATs can be listed as evaluation data, which introduces potential bias into our evaluation methodology. Second, due to the complex mechanisms of LLM-MAS, such as communication, memory, and reflection, the prompts are relatively long (approximately 2,000 tokens, with variation depending on the tokenizer used by the LLM). Additionally, process-oriented evaluation requires substantial interaction data, which leads to both low evaluation efficiency and significant token consumption, which is the common challenge across current methods for evaluating LLM-MAS capabilities. Third, the baseline used to evaluate LLM-MAS is composed of relatively simple structures, with the agent possessing only basic memory and reflection mechanisms, leaving substantial room for optimization.

\section*{Ethics Statement}

All human experiments were conducted with prior written informed consent from voluntary participants. Each participant was compensated fairly based on the duration of their engagement. No personally identifiable information was collected during the experiments. As our research focuses on collaboration in virtual environments, no foreseeable physical or psychological risks were posed to the participants.

\section*{Acknowledgments}
We thank the anonymous reviewers for their insightful comments. The research is supported by the Natural Science Foundation of Beijing, China (Grant No.L247010).

\newpage
\bibliography{acl_latex}

\newpage

\appendix

\section{Benchmark Detail}
\label{ap-benchmark-detail}

\subsection{Environment}
\label{ap-environment}

In this section, we provide a detailed overview of the Collab-Overcooked Benchmark environment design. We first introduce the interactive elements within the environment along with their layout. Next, we describe the action space available to agents. Finally, we present the methodology for defining layouts, enabling flexible modifications to the environment.

\subsubsection{Interactive Elements}

\begin{figure}[t]
  \includegraphics[width=\columnwidth]{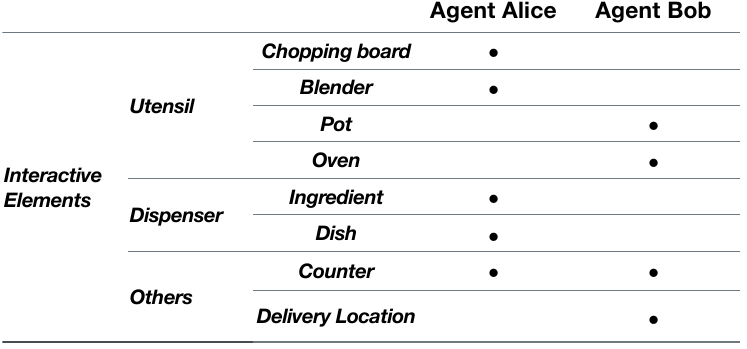}
  \caption{Interactive elements}
  \label{fig:appendix-interactive}
\end{figure}

Due to our resource isolation design, the interactive elements available to each agent differ. Figure \ref{fig:appendix-interactive} illustrates the interactive elements that both agents can engage with. We adopt the ``Forced Coordination'' level design from Overcooked-AI \cite{carroll2019utility}, where the two agents share only a single interactive element: the counter. This design necessitates resource exchange between agents to complete tasks.

We categorize interactive elements into three types: utensils, dispensers, and others. The details are as follows:
\begin{itemize} 
\item[$\bullet$] Utensils: These interactive elements take one or more ingredients as input and process them according to a predefined synthesis table, transforming them into new ingredients.  

\item[$\bullet$] Dispensers: Agents can retrieve ingredients or dishes from these elements, with the available items being predefined.  

\item[$\bullet$] Others: The counter serves as a critical interactive element for resource exchange between agents, allowing them to freely place or retrieve ingredients. The delivery location is where agents submit task outcomes. If the submitted ingredient meets the task requirements, the task is considered successful. Otherwise, incorrect submissions result in the removal of the submitted ingredient from the environment, often leading to task failure.  
\end{itemize}

\begin{table*}
  \centering
  \adjustbox{max width=\textwidth}{
\begin{tabular}{lcccccc}
\hline
\textbf{} & \textbf{Level 1} & \textbf{Level 2} & \textbf{Level 3} & \textbf{Level 4} & \textbf{Level 5} & \textbf{Level 6} \\ \hline
Average Recipe Token Count & 60.8 & 65.0 & 80.6 & 84.8 & 106.4 & 140.0 \\
Minimum Actions & 7 & 10 & 16 & 17 & 27 & 34 \\
Minimum Collaborative Actions & 2 & 5 & 7 & 9 & 14 & 19 \\
Interactive Elements Used & 4 & 5 & 7 & 6 & 8 & 8 \\ \hline
\end{tabular}}
\caption{\label{appendix-rebu-1}
    Statistics of recipe complexity across task levels, highlighting diversity in design, increasing difficulty, and interaction complexity}
\end{table*}

\subsubsection{Action Space}

The action space of each agent consists of a series of functions in the format ``func(args)'', which facilitate interactions with the environment or collaboration with other agents. Agent actions are categorized into shared actions and exclusive actions. Shared actions are common to both agents and include actions such as ``pickup'' (for picking up ingredients), ``place\_obj\_on\_counter'' (for interacting with the counter), ``put\_obj\_in\_utensil'' (for placing ingredients into utensils), and ``wait''. Exclusive actions, on the other hand, arise from the differing interactive elements in each agent’s environment. For example, Agent Bob has access to a pot, allowing it to perform the ``cook'' action, whereas Agent Alice, lacking a pot, cannot perform this action. Conversely, Agent Alice can interact with the chopping board to perform the ``cut'' action, which Agent Bob cannot. The specific actions available to Agent Alice and Agent Bob are listed as follows: 

\lstset{
    basicstyle=\ttfamily\small,  
    frame=single,               
    breaklines=true,             
    columns=fullflexible,        
    keepspaces=true,              
}

\label{lst-action-space}
\begin{lstlisting}[caption={Action Space List}]
Action Space for Agent Alice:
    1. pickup(obj,place)
    2. cut(chopping_board_name)
    3. stir(blender_name)
    4. place_obj_on_counter()
    5. put_obj_in_utensil(utensil)
    6. wait(num)   

Action Space for Agent Bob:
    1. pickup(obj,place)
    2. cook(pot_name)
    3. place_obj_on_counter()
    4. put_obj_in_utensil(utensil) 
    5. fill_dish_with_food(utensil)
    6. bake(oven_name)
    7. deliver()
    8. wait(num)
\end{lstlisting}

To accurately assess collaboration capabilities, we require that when an agent initiates collaboration, the initiating agent must encapsulate the desired action for the responding agent within a ``request''. This mechanism is utilized for calculating IC and RC. For example, if Agent Bob wants Agent Alice to retrieve an apple for it, Agent Bob will generate the following output: ``request(pickup(apple, ingredient\_dispenser)); request(place\_obj\_on\_counter())''. This request explicitly specifies the sequence of actions that Agent Alice is expected to execute, ensuring that the collaboration process is systematically coordinated.

\begin{table*}
  \centering
  \adjustbox{max width=\textwidth}{
\begin{tabular}{cccccc}
\hline
\multicolumn{1}{l}{\textbf{Complexity Level}} & \textbf{\begin{tabular}[c]{@{}c@{}}Acquiring\\ New Ingredients\end{tabular}} & \textbf{\begin{tabular}[c]{@{}c@{}}Processing the Ingredients \\ by Agent Alice\end{tabular}} & \textbf{\begin{tabular}[c]{@{}c@{}}Acquiring \\ a New Dish\end{tabular}} & \textbf{\begin{tabular}[c]{@{}c@{}}Processing the Ingredients \\ by Agent Bob\end{tabular}} & \textbf{\begin{tabular}[c]{@{}c@{}}Total Number of\\ Collaborative Actions\end{tabular}} \\ \hline
Level 1 & 1 & 0 & 0 & 1 & 2 \\
Level 2 & 1 & 1 & 1 & 1 & 5 \\
Level 3 & 1 & 1 & 1 & 2 & 7 \\
Level 4 & 2 & 1 & 1 & 2 & 9 \\
Level 5 & 2 & 2 & 1 & 3 & 12 \\
Level 6 & 3 & 3 & 1 & 4 & 17 \\ \hline
\end{tabular}}
\caption{\label{appendix-collaborative-actions}
    The number of collaborative behaviors under different complexity levels is given, as well as the total number of corresponding collaborative actions.}
\end{table*}

\subsubsection{Layout Definition Method}

We follow the environment design principles of Overcooked-AI \cite{carroll2019utility} and ProAgent \cite{zhang2024proagent}, enabling customization through external layout files. Compared to these prior works, our framework offers a broader range of configurable elements. For instance, the ``order\_probability'' parameter allows users to adjust the probability of tasks appearing randomly in the environment, while the ``recipes'' parameter enables customization of the synthesis list for each utensil. Further details can be found in the examples provided in our GitHub repository's layout files. Through our enhancements, nearly all aspects of the environment can be customized via a single external file, significantly enhancing the flexibility and scalability of our framework.

\subsection{Tasks Construction}
\label{ap-task-construction}

In this section, we provide detailed information about tasks, including task complexity level, task list, task recipe, and task RATs.

\subsubsection{Task complexity level}

Table~\ref{appendix-rebu-1} presents the statistics corresponding to different levels of task complexity. We have designed a series of task difficulty levels, ranging from basic ingredient transfer to complex recipe construction, requiring collaboration and error correction. The variation in external knowledge demands and environmental configurations substantially increases the challenges faced by LLM agents in terms of both comprehension and collaboration strategy formulation. Furthermore, we have incorporated additional interactive elements to expand the structural space of the tasks. The task levels demonstrate progressive increases in average recipe token count, minimum action requirements, collaboration frequency, and interaction complexity.

To characterize the complexity level of each task from the perspective of agent actions, we define four distinct types of collaborative behaviors. The complexity of a task is determined by the minimum number of such collaborative behaviors required for successful completion. The four categories of collaborative behaviors are defined as follows:

\begin{itemize} 

\item[$\bullet$] Acquiring New Ingredients: This behavior involves retrieving an ingredient from the Ingredient Dispenser. For example, Agent Alice might pick up an onion or an apple from the dispenser.

\item[$\bullet$] Processing the Ingredients: This behavior involves placing ingredients into a cooking utensil. For example, Agent Alice might place an ingredient on a chopping board or in a blender.

\item[$\bullet$] Acquiring a New Dish: This behavior involves retrieving a new dish from the Dish Dispenser. This action consists of a single step where Agent Alice picks up a dish.

\item[$\bullet$] Processing the Ingredients by Agent Bob: Similar to the first behavior, but performed by Agent Bob. This includes behaviors like placing an ingredient into a pot or an oven.
\end{itemize}

Each collaborative behavior corresponds to several collaborative actions. The complexity level of a task is calculated by summing the total number of collaborative actions required from each behavior. Specifically, the number of actions in each of the four categories is counted based on the task's requirements. This approach ensures that tasks with more complex or numerous collaboration requirements are considered more difficult than those with fewer actions. Table \ref{appendix-collaborative-actions} provides statistical data on collaborative behaviors and collaborative actions.

Each task's RATs provide the exact number of actions for each type of collaboration, which is used to determine the total complexity level for that task. The complexity calculation allows for a comparison of tasks, ensuring that they are evaluated based on their collaborative complexity.

\subsubsection{Task List}

Table \ref{tab:appendix-task-name} presents a list of task names across 6 complexity levels, comprising a total of 30 tasks. As indicated by the task names, tasks within the same complexity level share identical workflows, with the only variation being the selection of ingredients. This design aims to mitigate potential biases in LLMs towards specific ingredients, thereby reducing evaluation discrepancies caused by such biases.

\subsubsection{Recipes}

Each task corresponds to a recipe that outlines the workflow required to complete the task, including the necessary ingredients and cooking steps. There are two important aspects to note regarding the recipe: First, one cooking step typically involves multiple actions by the agents. This necessitates that the agents carefully decompose the cooking step into specific actions after thoroughly understanding both the recipe and the environment. Second, some cooking steps can be executed in a different order. For instance, when multiple ingredients require pre-processing, followed by combining the processed ingredients into a utensil for further preparation, the order in which the ingredients are preprocessed can be interchanged. This decision is typically made by the agents, leading to the possibility of multiple valid RATs for the same task. Allowing such flexibility is both reasonable and aligned with real-world practices. Listing \ref{lst-recipe} is an example of the recipe for ``Baked Pumpkin Soup'', which includes the recipe name, required ingredients with quantities, and detailed cooking instructions.

\newpage

\begin{lstlisting}[caption={Recipe example},label={lst-recipe}]
NAME:
Baked Pumpkin Soup

INGREDIENTS:
pumpkin(1)

COOKING STEPs:
1. Cut a pumpkin into slices.
2. Place the pumpkin slices in the oven and bake for 3 timesteps.
3. Transfer the baked pumpkin slices to a pot and cook for 3 timesteps.
4. Fill a dish with the soup from the pot and deliver.
\end{lstlisting}

\begin{lstlisting}[caption={RAT of "Baked Pumpkin Soup" task},label={lst-rat}]
"RAT_1": 
{
    "agent_0": [
        "pickup(pumpkin_slices, counter)",
        "put_obj_in_utensil(oven0)",
        "bake(oven0)",
        "pickup(baked_pumpkin_slices, oven0)",
        "put_obj_in_utensil(pot0)",
        "cook(pot0)",
        "pickup(dish,counter)",
        "fill_dish_with_food(pot0)",
        "deliver()"
    ],
    "agent_1": [
        "pickup(pumpkin, ingredient_dispenser)",
        "put_obj_in_utensil(chopping_board0)",
        "cut(chopping_board0)",
        "pickup(pumpkin_slices,chopping_board0)",
        "place_obj_on_counter()",
        "pickup(dish,dish_dispenser)",
        "place_obj_on_counter()"
    ]
}
\end{lstlisting}

\subsubsection{Referential Action Trajectory}

To evaluate the agents' collaboration capabilities both in terms of end-to-end and process-oriented metrics, we provide the RATs for each task. Given that our tasks are sequential process-specific, we assume that the RATs can be exhaustively enumerated or largely known. We have annotated the RATs for each task, which include the optimal referential action sequences for both agents to complete the task. Each RAT ensures that the agents can accomplish the task with a minimal number of actions, while also employing the optimal strategy to parallelize certain actions for efficiency. A task may have multiple valid RATs, for example, the order in which two ingredients are retrieved may not affect the overall task completion time. During evaluation, the TES and ITES functions select the RAT with the highest matching score as the reference for assessment. Listing \ref{lst-rat} provides an example of the RATs for the ``Baked Pumpkin Soup'' task, with separate RATs for each of the two agents. Because the ``Baked Pumpkin Soup'' task has only one completed route, there is only one RAT.

\begin{figure*}[t]
  \includegraphics[width=1\linewidth]{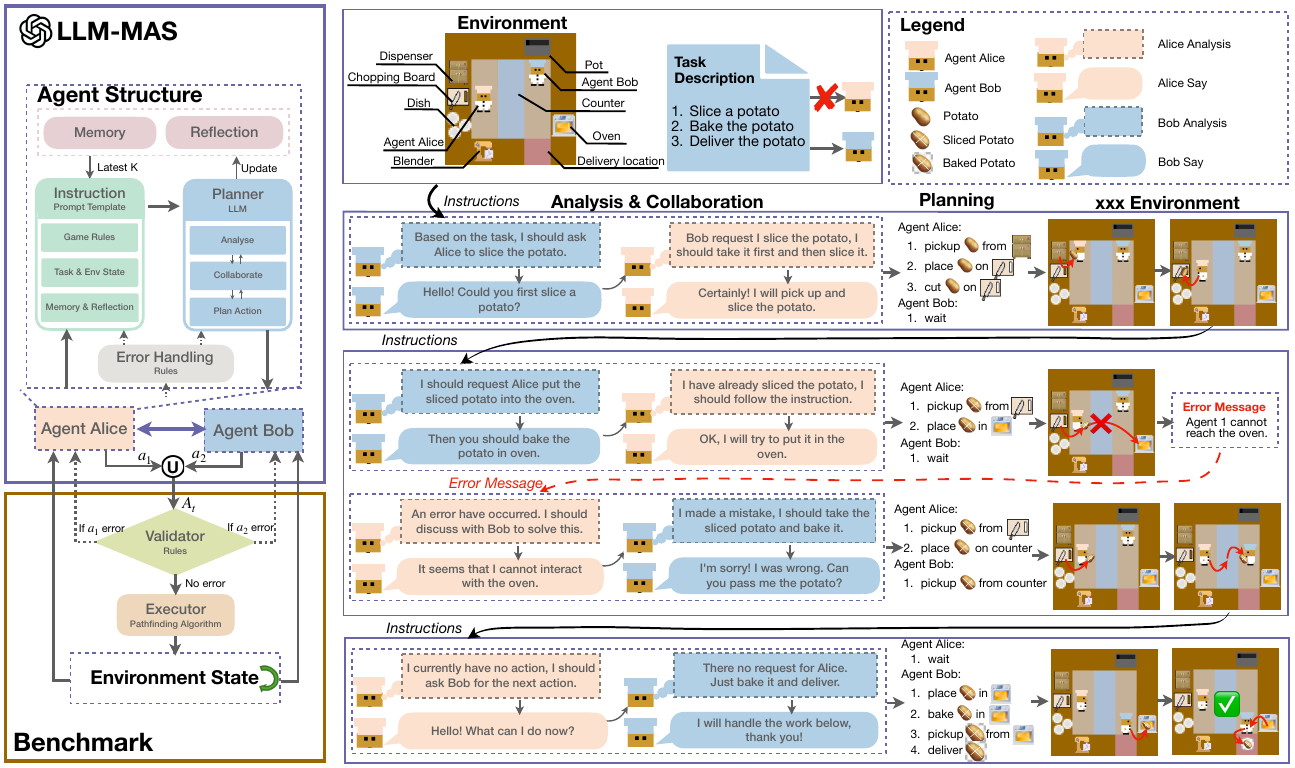}
  \caption {The left side of the figure presents the baseline architecture used for evaluating different LLMs, where Agent Alice and Agent Bob share the same structural design, differing only in their prompt. The right side of the figure illustrates the interaction process between the two agents as they collaborate to complete the ``Baked Potato Slices'' task within our benchmark. This includes the agents' analytical processes as well as a record of their natural language communication.}
  \label{fig:appendix-baseline}
\end{figure*}

\begin{table}
  \centering
\begin{tabular}{cc}
\hline
\begin{tabular}[c]{@{}c@{}}Complexity\\ Level\end{tabular} & Task Name \\ \hline
\multirow{5}{*}{Level 1} & Baked Bell Pepper \\ \cline{2-2} 
 & Baked Sweet Potato \\ \cline{2-2} 
 & Boiled Egg \\ \cline{2-2} 
 & Boiled Mushroom \\ \cline{2-2} 
 & Boiled Sweet Potato \\ \hline
\multirow{5}{*}{Level 2} & Baked Potato Slices \\ \cline{2-2} 
 & Baked Pumpkin Slices \\ \cline{2-2} 
 & Boiled Corn Slices \\ \cline{2-2} 
 & Boiled Green Bean Slices \\ \cline{2-2} 
 & Boiled Potato Slices \\ \hline
\multirow{5}{*}{Level 3} & Baked Bell Pepper Soup \\ \cline{2-2} 
 & Baked Carrot Soup \\ \cline{2-2} 
 & Baked Mushroom Soup \\ \cline{2-2} 
 & Baked Potato Soup \\ \cline{2-2} 
 & Baked Pumpkin Soup \\ \hline
 & \begin{tabular}[c]{@{}c@{}}Sliced Bell Pepper\\ and Corn Stew\end{tabular} \\ \cline{2-2} 
 & \begin{tabular}[c]{@{}c@{}}Sliced Bell Pepper\\ and Lentil Stew\end{tabular} \\ \cline{2-2} 
Level 4 & \begin{tabular}[c]{@{}c@{}}Sliced Eggplant\\ and Chickpea Stew\end{tabular} \\ \cline{2-2} 
 & \begin{tabular}[c]{@{}c@{}}Sliced Pumpkin\\ and Chickpea Stew\end{tabular} \\ \cline{2-2} 
 & \begin{tabular}[c]{@{}c@{}}Sliced Zucchini\\ and Chickpea Stew\end{tabular} \\ \hline
 & \begin{tabular}[c]{@{}c@{}}Mashed Broccoli\\ and Bean Patty\end{tabular} \\ \cline{2-2} 
 & \begin{tabular}[c]{@{}c@{}}Mashed Carrot\\ and Chickpea Patty\end{tabular} \\ \cline{2-2} 
Level 5 & \begin{tabular}[c]{@{}c@{}}Mashed Cauliflower\\ and Lentil Patty\end{tabular} \\ \cline{2-2} 
 & \begin{tabular}[c]{@{}c@{}}Mashed Potato\\ and Pea Patty\end{tabular} \\ \cline{2-2} 
 & \begin{tabular}[c]{@{}c@{}}Mashed Sweet Potato\\ and Bean Patty\end{tabular} \\ \hline
 & \begin{tabular}[c]{@{}c@{}}Potato Carrot\\ and Onion Patty\end{tabular} \\ \cline{2-2} 
 & \begin{tabular}[c]{@{}c@{}}Romaine Lettuce Pea\\ and Tomato Patty\end{tabular} \\ \cline{2-2} 
Level 6 & \begin{tabular}[c]{@{}c@{}}Sweet Potato Spinach\\ and Mushroom Patty\end{tabular} \\ \cline{2-2} 
 & \begin{tabular}[c]{@{}c@{}}Taro Bean\\ and Bell Pepper Patty\end{tabular} \\ \cline{2-2} 
 & \begin{tabular}[c]{@{}c@{}}Zucchini Green Pea\\ and Onion Patty\end{tabular} \\ \hline
\end{tabular}
  \caption{The names of 30 tasks in total are divided into 6 complexity levels.}
  \label{tab:appendix-task-name}
\end{table}

\subsection{Baseline}

\label{ap-baseline}

In this section, we introduce the baseline structure and prompt design we use to test different LLMs.

\subsubsection{Baseline Construction}

Figure \ref{fig:appendix-baseline} illustrates the structure of the baseline and provides an example of agents interacting and collaborating to complete a task within our benchmark. The baseline architecture consists of an Instruction-Builder, Planner, Communication, Error-Handling, Memory, and Reflection modules. The structure remains identical across different agents, with variations arising only in the environment descriptions, action spaces, and task-specific knowledge provided within the prompts.

\paragraph{Instruction-builder}

The Instruction-builder is a rule-based module responsible for managing and integrating the prompts for each agent. It reads the state dictionary from the environment and fills in a prompt template. The prompt template includes both fixed prompts and slot-based prompts. Fixed prompts contain: (1) game rules, such as objectives, scoring workflows, functions of each kitchen utensils, and methods for preparing dishes; (2) communication rules and output format specifications; and (3) a definition of the agent’s action space, along with a brief description of actions available to teammates. Slot-based prompts include: (1) the current recipe for the task (if the agent has access to the recipe); (2) the current environment observations, such as kitchen layout and teammate status; (3) communication records with other agents up to the current time step; and (4) memory and reflection from previous time steps.

\paragraph{Planner}

The planner is the core decision-making component for the agent. It generates three fields: ``Analysis'', ``Say'', and ``Plan''. The ``Analysis'' field represents the agent’s assessment of the current environment state, task, and memories, assisting the planner in making informed decisions. The ``Say'' field determines whether collaboration is required; if the planner identifies a need for collaboration, it generates communication content directly in this field. The ``Plan'' field contains the action sequence that the planner has devised for the agent.

\paragraph{Communication}

Communication between agents enables the transmission of collaborative intentions or requests for assistance. When communication content is detected in the ``Say'' field, all agents enter the communication channel. Within this channel, each agent speaks in sequence until a special token ``[END]'' is generated or the maximum number of interaction rounds is reached. Once communication is complete, agents formulate their plans based on the information exchanged.

\paragraph{Error-handling}

The error-handling process manages situations in which the generated actions are deemed invalid by the environment. When an agent receives an error message from the environment, the error information is incorporated into the prompt and re-entered into the planner. This cycle continues until the generated actions are considered valid by the environment or the maximum number of attempts is reached.

\paragraph{Memory and Reflection}

Memory and reflection represent the accumulation of an agent’s past experiences, enabling it to engage in long-term planning. We implement memory and reflection using a straightforward approach. The memory logs the action sequences that the agent has completed in the past, while the reflection records the previous agent’s reflections on invalid actions.

\subsubsection{Prompt}

In this section, we provide a detailed description of the prompts used to drive LLM-based agents. Since LLM-MAS involves multiple agents interacting within an environment, the prompt design is inherently more complex than that of a single-agent system. Each request to the LLM typically consumes approximately 2,000 tokens, with slight variations depending on the specific tokenizer used by the LLM. To structure this complexity, we categorize the prompts into three key components: game rules, action space definitions, and input-output format specifications. We will elaborate on each component and provide illustrative examples to demonstrate their implementation.

\paragraph{Game Rules}

\begin{figure*}[t]
  \includegraphics[width=1\linewidth]{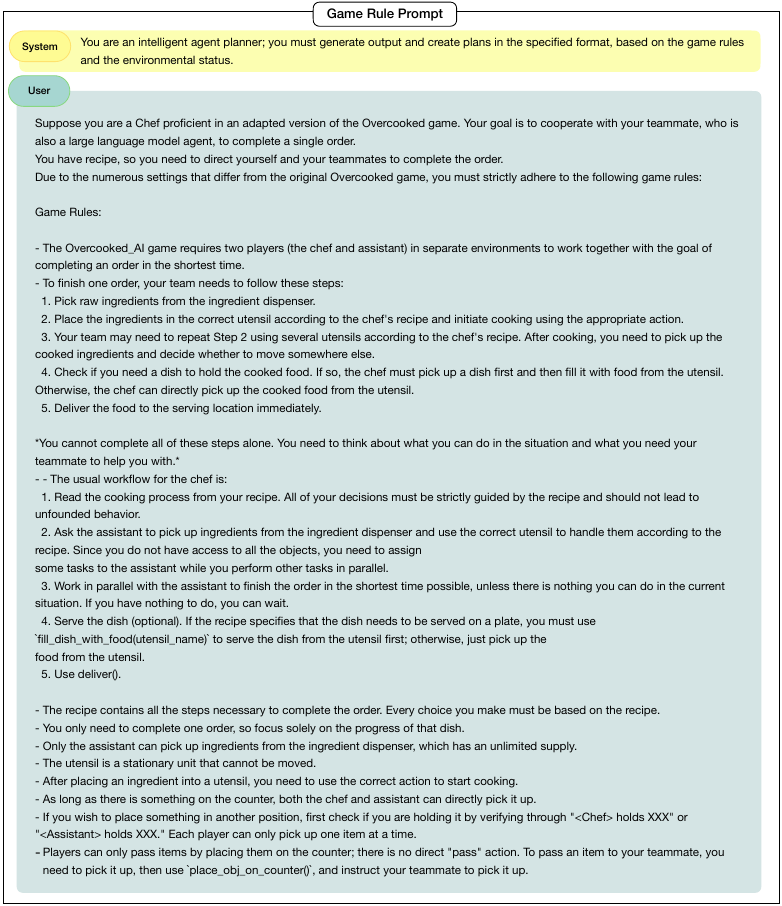}
  \caption {Prompt for game rules.}
  \label{appendix:game-rule}
\end{figure*}

The game rules part of the prompt defines the task objective, agent roles, and interaction constraints. It outlines the step-by-step workflow for completing an order, emphasizing task division, coordination, and strict adherence to recipe instructions. Figure \ref{appendix:game-rule} shows all the content of the game rule prompt.

\paragraph{Action Space Definitions}

\begin{figure*}[t]
  \includegraphics[width=1\linewidth]{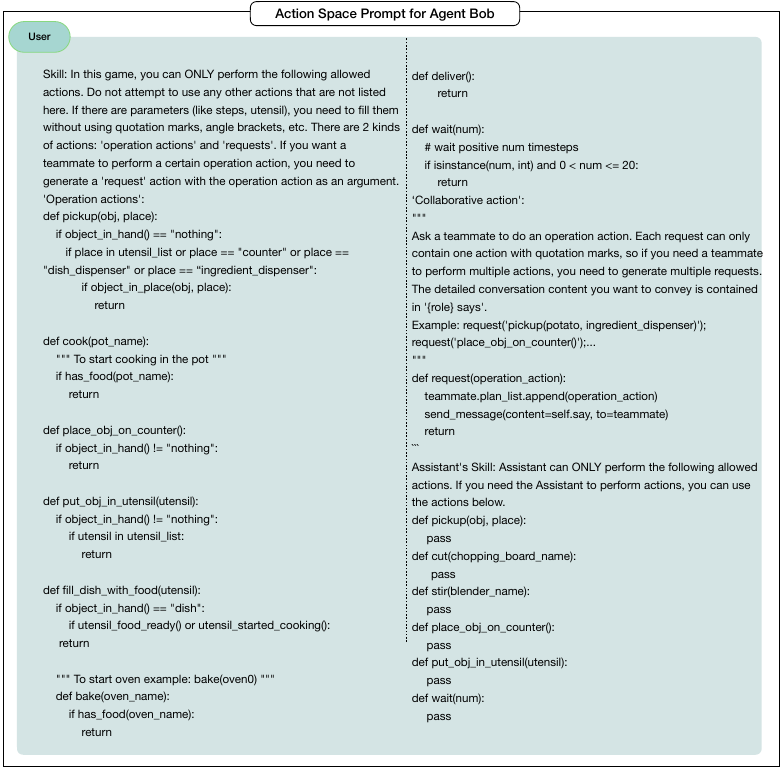}
  \caption {Prompt for the action space of Agent Bob.}
  \label{appendix:action-bob}
\end{figure*}

This part of the prompt defines the action space for Agent Bob, following the action specification method used in ProAgent \cite{zhang2024proagent}. It categorizes actions into operation actions (directly executable by the agent) and collaborative actions (requests for the teammate to perform an action). Figure \ref{appendix:action-bob} shows the prompt of Agent Bob's action space.

\paragraph{Input-Output Format}

\begin{figure*}[t]
  \includegraphics[width=1\linewidth]{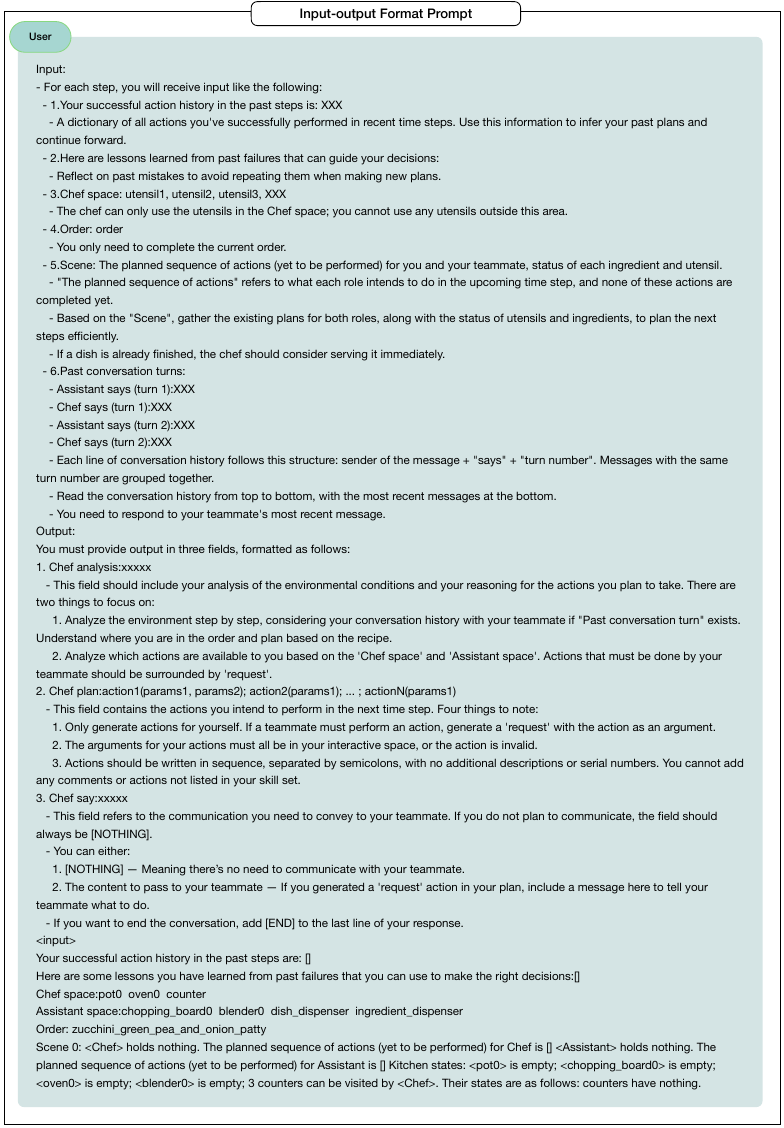}
  \caption {Prompt for the input-output format.}
  \label{appendix:io-format}
\end{figure*}

The input-output format part defines the structured information provided to the agent at each step and the required response format. The input includes past action history, lessons from failures, available utensils, the current order, the planned sequence of actions, and past conversations. The output consists of three fields: analysis (environment assessment and reasoning for actions), plan (the agent’s planned actions for the next step), and say (communication with the teammate, if necessary). This structured format ensures that the agent can make informed decisions, coordinate effectively, and execute tasks systematically. \ref{appendix:io-format} shows all the content of the input-output format prompt.

The above section outlines the key prompts used to drive the LLM agents. For further details regarding prompts related to memory, reflection, and other components, please refer to the comprehensive prompts provided in our GitHub repository.

\section{Evaluation}

\subsection{Details in TES}

\label{ap-sea}

The TES is formally expressed as:  

\begin{equation}
\text{TES}(\overline{h}_k) = \max_j\left\{\frac{(1+\beta^2)D_{\text{max}}^j(\overline{h}_k, \overline{g}_k^j)}{m_k + \beta^2 n_k}\right\}
\end{equation}

\noindent where $\overline{h}_k= \{a_k^1, a_k^2, \ldots, a_k^T\}$ is the historical action sequence up to timestep $T$ of agent $k$, $\overline{g}_k^j=\{g_i\}_{i=1}^{m_k} \in \mathcal{R}$ is $j$-th RAT of agent $k$, $\beta$ is the hyperparameter balancing the weight of task progress and redundancy, and $D_{\text{max}}^j(\overline{h}_k, \overline{g}_k^j)$ computes the length of the longest order-preserving subsequence in $\overline{h}_k$ that matches $\overline{g}_k^j$:  

\begin{multline}
     D_{max}^j = \max_d \{ d \mid \forall \, 1 \leq i_1 < \dots < i_d \leq n_k, \, \\
     \text{s.t.} \ a_{i_1} = g_1, a_{i_2} = g_2, \dots, a_{i_d} = g_d \}
\end{multline}

It is important to note that the TES function introduces modifications to the Longest Common Subsequence (LCS) calculation in ROUGE-L \cite{lin2004rouge}. These modifications are driven by one main reason: Improved identification of redundant actions. Listing \ref{lst-example} illustrates a very common scenario where, due to the agent’s incorrect choice in step four, the fifth step fails to advance the task. Specifically, the agent places an irrelevant item, ``egg'', onto the counter, which does not contribute to the task's progress. In this case, the standard ROUGE-L, based on LCS, would mistakenly consider the agent's fifth action as matching the RAT, leading to an inflated evaluation score.

TES overcomes this limitation by combining maximal order-preserving alignment with efficiency-aware normalization, making it well-suited for collaborative tasks requiring synchronized, sequence-specific interactions.

\begin{lstlisting}[caption={Comparison of TES with other functions},label={lst-example}]
Example:
RAT: 
    1. pickup(tofu, ingredient_dispenser)
    2. put_obj_in_utensil(chopping_board_0)
    3. cut(chopping_board_0)
    4. pickup(chopped_tofu, chopping_board_0)
    5. place_obj_on_counter()
Agent Action Trajectory: 
    1. pickup(tofu, ingredient_dispenser)
    2. put_obj_in_utensil(chopping_board_0)
    3. cut(chopping_board_0)
    4. pickup(egg, ingredient_dispenser)
    5. place_obj_on_counter()
Result:
    ROUGE-L: 0.8
    TES: 0.6
\end{lstlisting}

\subsection{Details in IC and RC}

Initiating Capability (IC) and Responding Capability (RC) are proposed to evaluate the LLM agent's capabilities to initiate and respond to collaboration, respectively. Physically, these metrics represent the success rate of an LLM agent in initiating or responding to collaborative behaviors within a given task. The determination of success is based on the change in ITES induced by the newly proposed action compared to historical actions. Taking collaboration initiation as an example, a newly initiated collaborative action $a$ is considered successful if it results in an increase in ITES, i.e., $\text{ITES} > 0$. This reflects whether the proposed action $a$ contributes to the advancement of the task; if so, it is deemed a successful collaboration attempt. This evaluation paradigm has been widely adopted in prior research \cite{gong2023mindagent,hong2023metagpt,mandi2024roco}, and thus, both IC and RC are not only grounded in meaningful physical interpretations but also serve as effective indicators of real-world collaborative performance.

\section{Supplementary Experiment}

In this section, we present supplementary experiments that support the conclusions of the main body. First, we investigate the impact of different hyperparameter values for $\gamma$ on the task completion success rate of the LLM-MAS and provide the rationale for selecting $\gamma = 1.5$. Next, we describe the details of the human performance evaluation, including the experimental design and the human-computer interaction interface. Additionally, we introduce new recipes and additional results presented in the failure analysis section. Finally, we provide case studies illustrating both successful and unsuccessful task completions by the LLM-MAS.

\subsection{Impact of Varying \texorpdfstring{$\gamma$}{gamma} on Task Success Rate}

\label{ap-gamma}

The hyperparameter $\gamma$ controls the task failure threshold. Specifically, it determines a time constraint on the task, which is calculated by multiplying the optimal completion time by the value of $\gamma$. As $\gamma$ increases, the task success rate (SR) of the LLM-MAS will improve, as the system is allowed more time to complete the task. However, $\gamma$ cannot be increased indefinitely, as doing so would lead to inefficiencies in the evaluation process. An excessively high value of $\gamma$ might artificially inflate the success rate, as the extended time window may not reflect the true capabilities of the model in real-world scenarios, and it wastes computing resources. On the other hand, setting $\gamma$ too low could result in an overly strict evaluation, where the system is unable to complete tasks even when it could have more time. Therefore, it is essential to select an optimal value for $\gamma$ that balances both task success and evaluation efficiency. 

\begin{figure*}[t]
  \includegraphics[width=1\linewidth]{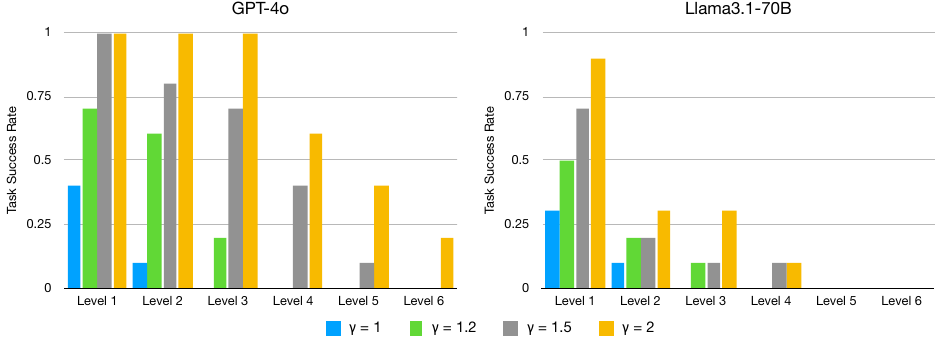}
  \caption {The task success rates of the GPT-4o and Llama-3.1-70B  at 6 complexity levels under different $\gamma$ values.}
  \label{appendix-gamma}
\end{figure*}

\begin{table*}
  \centering
  \adjustbox{max width=\textwidth}{
\begin{tabular}{l|cccc|cccc|cccc}
\hline
\multicolumn{1}{c|}{} & \multicolumn{4}{c|}{\textbf{Level 1}} & \multicolumn{4}{c|}{\textbf{Level 2}} & \multicolumn{4}{c}{\textbf{Level 3}} \\
\multicolumn{1}{c|}{} & \textbf{SR} & \textbf{PC} & \textbf{IC} & \textbf{RC} & \textbf{SR} & \textbf{PC} & \textbf{IC} & \textbf{RC} & \textbf{SR} & \textbf{PC} & \textbf{IC} & \textbf{RC} \\ \hline
\textbf{10s} & \multicolumn{1}{l}{100.0} & \multicolumn{1}{l}{96.8} & \multicolumn{1}{l}{87.1} & \multicolumn{1}{l|}{90.0} & 100.0 & 94.2 & 85.3 & 90.3 & 90.0 & 91.6 & 85.8 & 89.1 \\
\textbf{15s} & \multicolumn{1}{l}{100.0} & \multicolumn{1}{l}{96.3} & \multicolumn{1}{l}{90.9} & \multicolumn{1}{l|}{91.4} & 100.0 & 96.0 & 89.2 & 90.1 & 100.0 & 94.7 & 86.4 & 87.2 \\
\textbf{20s} & \multicolumn{1}{l}{100.0} & \multicolumn{1}{l}{97.9} & \multicolumn{1}{l}{98.0} & \multicolumn{1}{l|}{98.0} & 100.0 & 98.9 & 96.1 & 97.0 & 100.0 & 99.4 & 93.0 & 94.0 \\ \hline
 & \multicolumn{4}{c|}{\textbf{Level 4}} & \multicolumn{4}{c|}{\textbf{Level 5}} & \multicolumn{4}{c}{\textbf{Level 6}} \\
 & \textbf{SR} & \textbf{PC} & \textbf{IC} & \textbf{RC} & \textbf{SR} & \textbf{PC} & \textbf{IC} & \textbf{RC} & \textbf{SR} & \textbf{PC} & \textbf{IC} & \textbf{RC} \\ \hline
\textbf{10s} & 100.0 & 93.2 & 86.4 & 86.2 & 90.0 & 88.5 & 80.3 & 82.5 & 90.0 & 85.6 & 78.3 & 82.1 \\
\textbf{15s} & 90.0 & 94.1 & 87.0 & 87.6 & 100.0 & 90.7 & 84.8 & 86.4 & 90.0 & 91.5 & 80.9 & 83.6 \\
\textbf{20s} & 100.0 & 95.8 & 93.5 & 94.5 & 90.0 & 96.6 & 91.5 & 93.0 & 100.0 & 95.1 & 87.5 & 90.5 \\ \hline
\end{tabular}}
\caption{\label{appendix-human-performance}
    Human performance across 6 complexity levels under different time constraints (10s, 15s, and 20s per step), where participants were required to complete communication, reasoning, and action selection within the allotted duration at each time step.}
\end{table*}

Figure \ref{appendix-gamma} illustrates the task success rates of GPT-4o and Llama-70B at 6 complexity levels under varying values of the hyperparameter $\gamma$. We observed that when $\gamma = 1$, which requires completing tasks along the optimal path, even a highly capable model like GPT-4o failed to complete the majority of tasks. However, when $\gamma$ was increased to 1.5 or 2, GPT-4o was able to complete most tasks at complexity levels 4 and below. We chose $\gamma = 1.5$ rather than $\gamma = 2$ because, for models with fewer parameters than GPT-4o, such as Llama-3.1-70B, increasing $\gamma$ does not significantly improve success rates on higher complexity tasks. In fact, most models we tested struggled to complete tasks above level 4, often requiring the maximum time limit during evaluations. By selecting $\gamma = 1.5$, we were able to save approximately 33\% of computational resources compared to using $\gamma = 2$, thereby enabling a more efficient evaluation of the LLM’s capabilities.

\subsection{Human Performance Evaluation}

\label{ap-human}

\subsubsection{Experiment Setup}

\begin{figure*}[t]
  \includegraphics[width=1\linewidth]{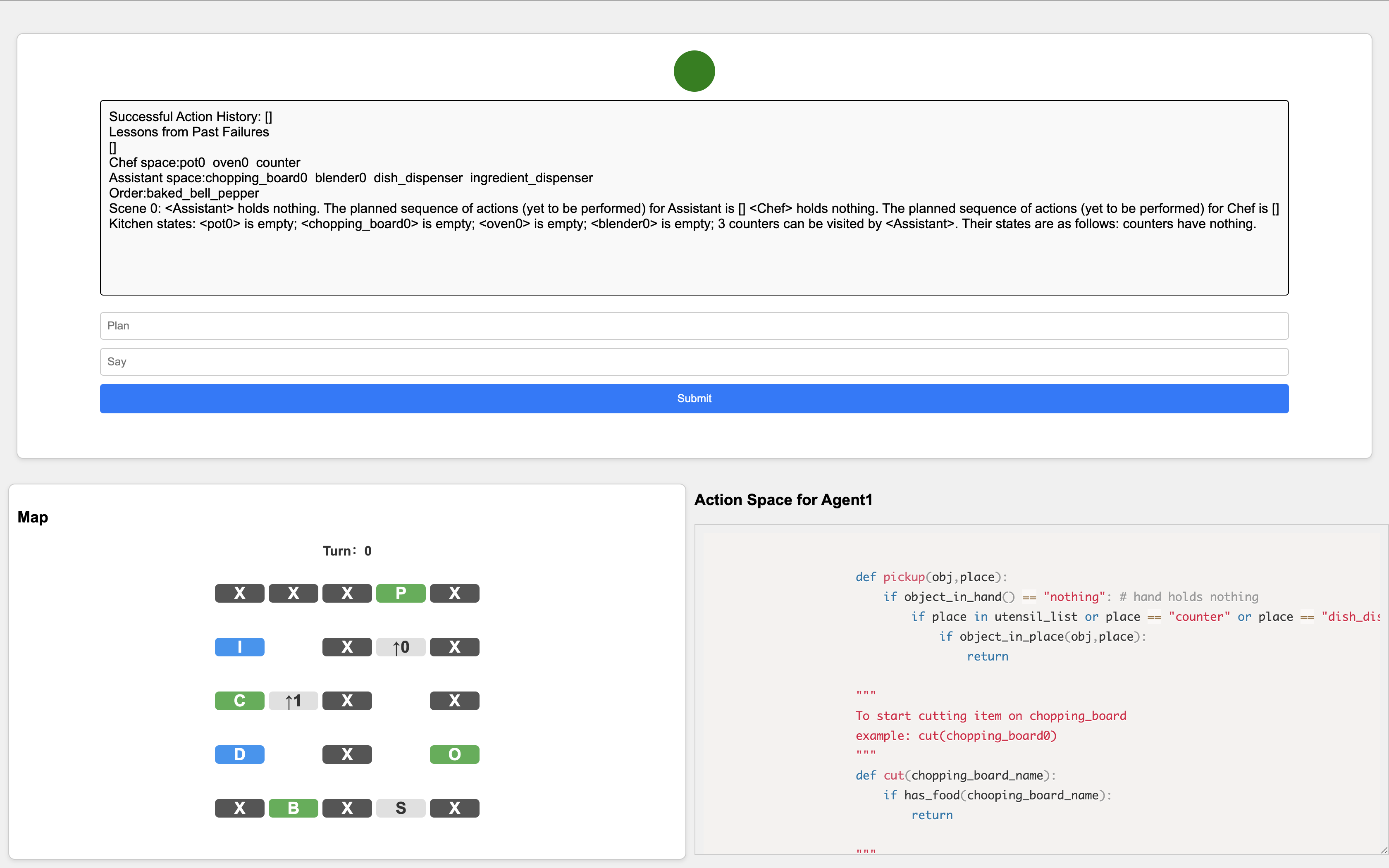}
  \caption {Human-computer interaction as Agent Alice.}
  \label{appendix:human-alice}
\end{figure*}

\begin{figure*}[t]
  \includegraphics[width=1\linewidth]{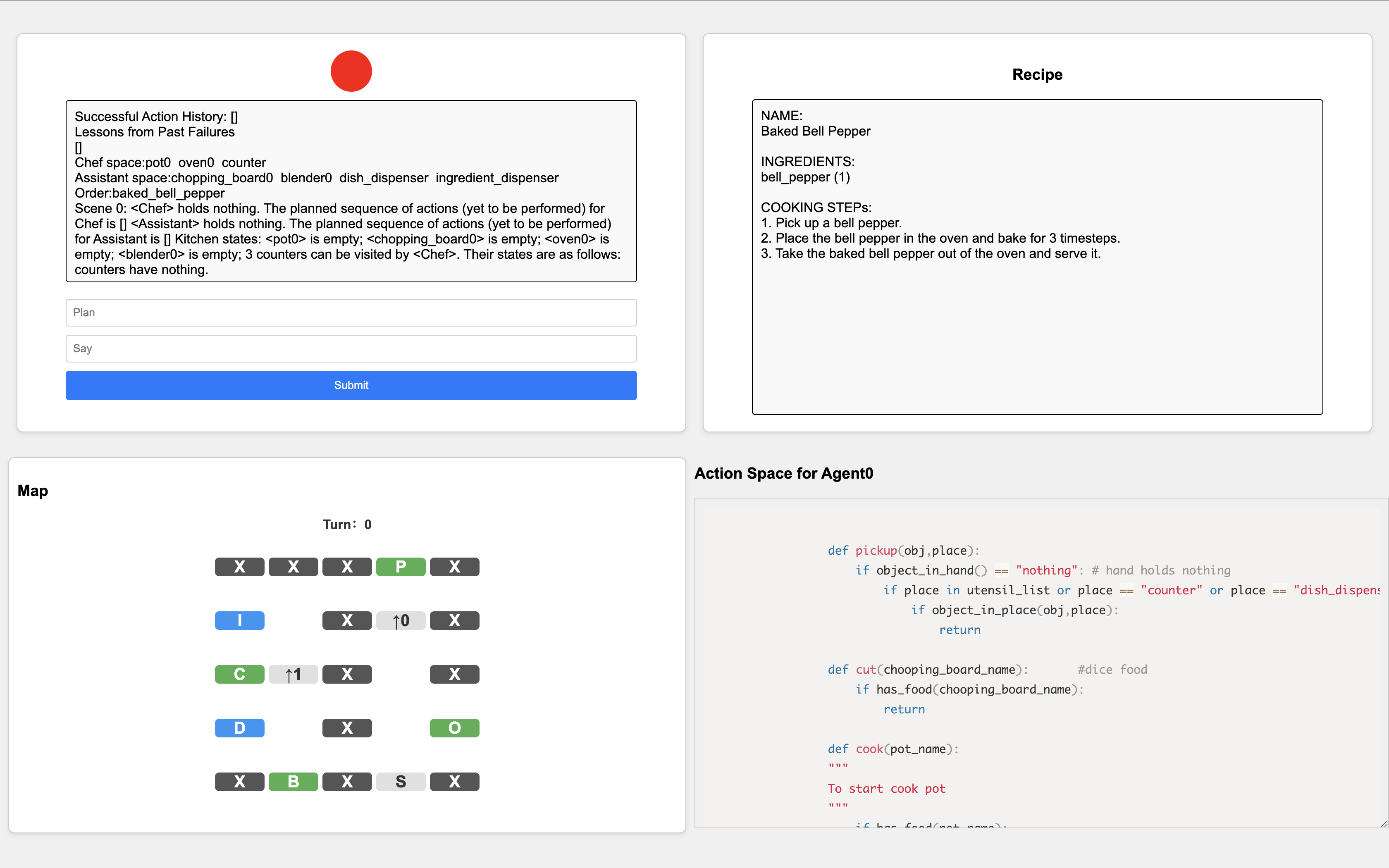}
  \caption {Human-computer interaction as Agent Bob.}
  \label{appendix:human-bob}
\end{figure*}

To evaluate human performance on our benchmark, we invited ten volunteers to participate in our experiments. The participants were organized into five pairs, with each pair assigned two randomly selected tasks from each complexity level. Consequently, each complexity level was tested ten times. To ensure participants fully understood the game rules, the available action space, the input-output format, and the current state of the environment, we designed a dedicated human-computer interaction interface. This interface presented the prompts originally inputted to the agent in a human-friendly format, without revealing any additional information beyond what was accessible to the agent. Figures \ref{appendix:human-alice} and \ref{appendix:human-bob} illustrate the layout of this interface.

To further regulate the decision-making process and assess human performance under time-constrained conditions, we imposed temporal limits on each decision step. Specifically, participants were instructed to complete their communication, reasoning, and action selection within a total duration of 10, 15, or 20 seconds per time step. Each of these time limits was evaluated across trials to investigate their effects. The action was considered successfully generated if the participant verbally expressed their intended move before the time expired. The subsequent process of inputting the action into the environment was excluded from the timing. Moreover, unlike previous implementations that required typed communication, participants in this experiment were permitted to communicate verbally, thereby enhancing the naturalness and efficiency of interaction.

\subsubsection{Discussion}
Table~\ref{appendix-human-performance} presents the performance of human participants under varying time constraints imposed on communication, reasoning, and action selection. Although these constraints led to a measurable decline in performance, human participants consistently achieved comparable performance across tasks of different complexity levels. In terms of end-to-end metrics, including SR and PC, the performance degradation was primarily reflected in an increased number of redundant actions, which resulted in a lower PC. However, SR remained relatively stable, as participants were generally able to recover quickly from suboptimal decisions. Regarding process-oriented metrics, such as IC and RC, human performance showed minimal discrepancy between IC and RC, suggesting a balanced ability to both initiate and respond in collaborative contexts. In contrast, LLM-based agents exhibited a more pronounced gap between IC and RC, consistent with prior findings that highlight their difficulty in initiating collaboration \cite{li2023theory}. These results indicate that in the Collab-Overcooked environment, humans are able to decompose and allocate tasks with relative ease, whereas LLMs face substantial challenges in doing so.

\subsection{Supplement to Correlation Analysis}
\label{appendix-corr-supp}

Section~\ref{fa} presented our analysis of attention distribution differences under successful and failed collaboration scenarios, from the perspective of model behavior. In this section, we provide additional experimental details and present more comprehensive results.

\begin{figure}[t]
  \includegraphics[width=1\linewidth]{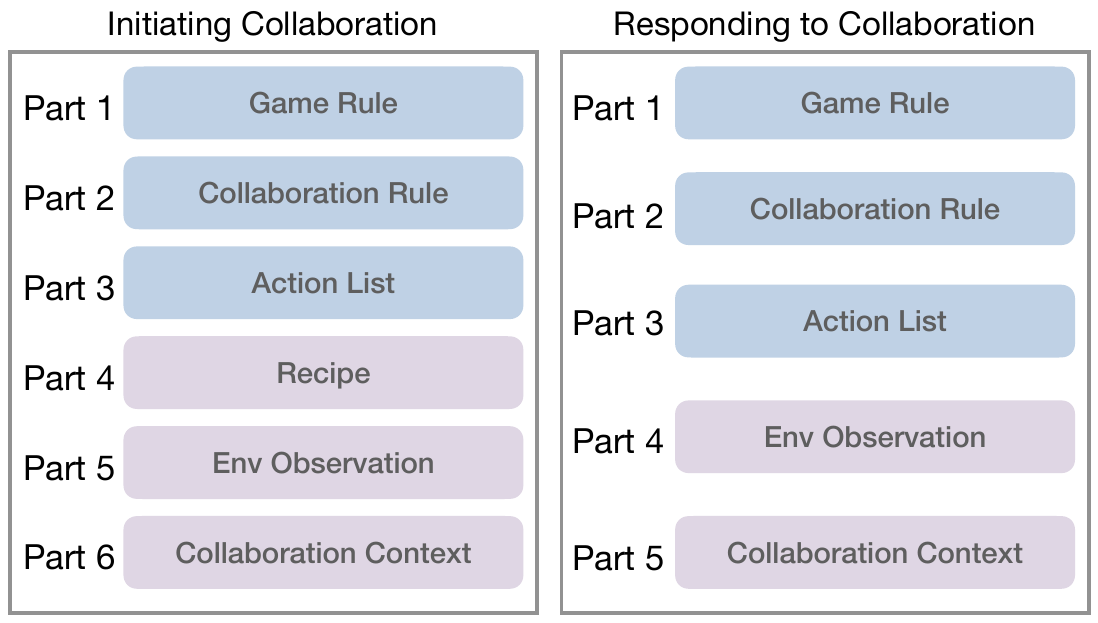}
  \caption {Segmentation of prompt components provided to agents at different stages. Notably, recipe information is omitted during response to collaboration, establishing an asymmetry in task-relevant input. The ``Collaboration Context'' encodes both prior interactions and the current collaboration instruction}
  \label{appendix-prompt-part}
\end{figure}

As shown in Figure~\ref{appendix-prompt-part}, the prompt provided to LLM-MAS is segmented into five or six parts, depending on whether the agent is initiating or responding to collaboration. To compute the attention distribution, we measure the cumulative attention assigned by the model's first generated token to each prompt part. We then compare these distributions between successful and failed collaboration cases.

\begin{figure*}[t] 
    \centering
    \begin{subcaptionbox}{Initiating Collaboration\label{appendix-corr-init}}{
        \includegraphics[width=0.9\linewidth]{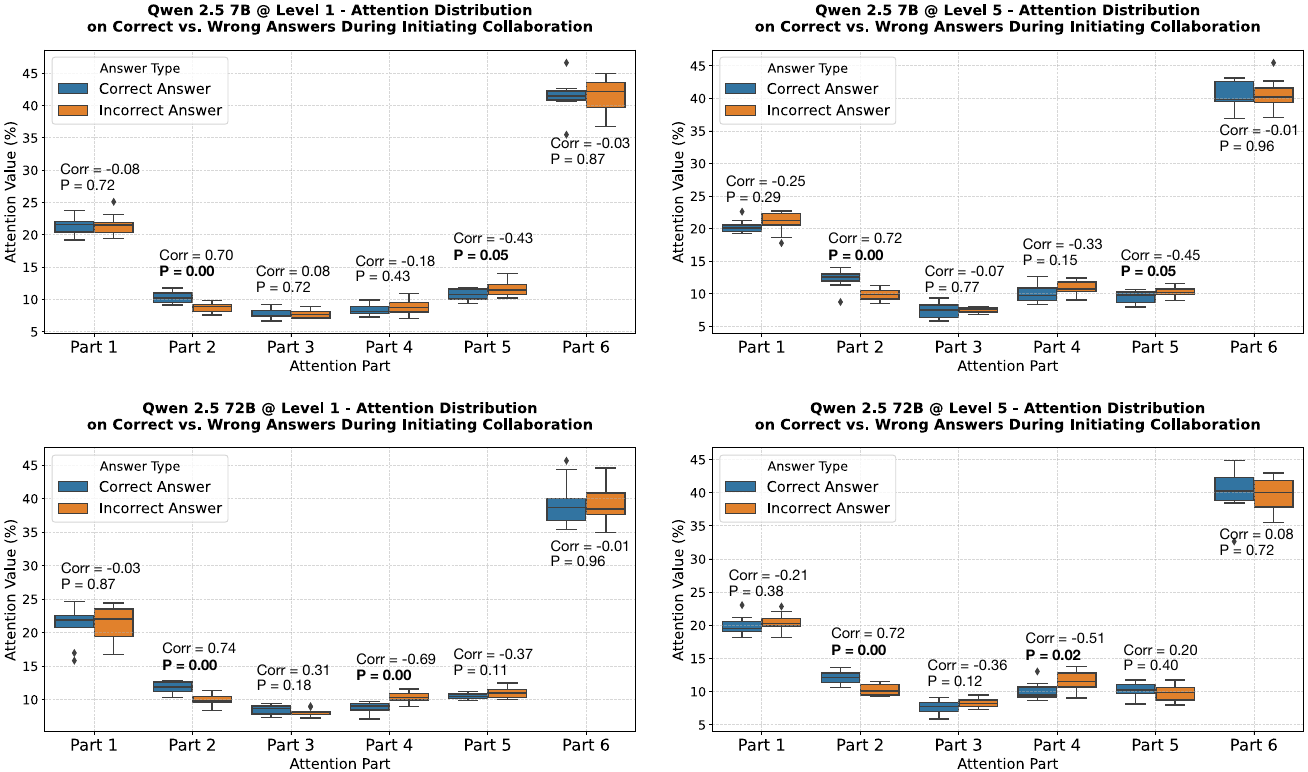}
    }
    \end{subcaptionbox}

    \vspace{1em} 

    \begin{subcaptionbox}{Responding to Collaboration\label{appendix-corr-resp}}{
        \includegraphics[width=0.9\linewidth]{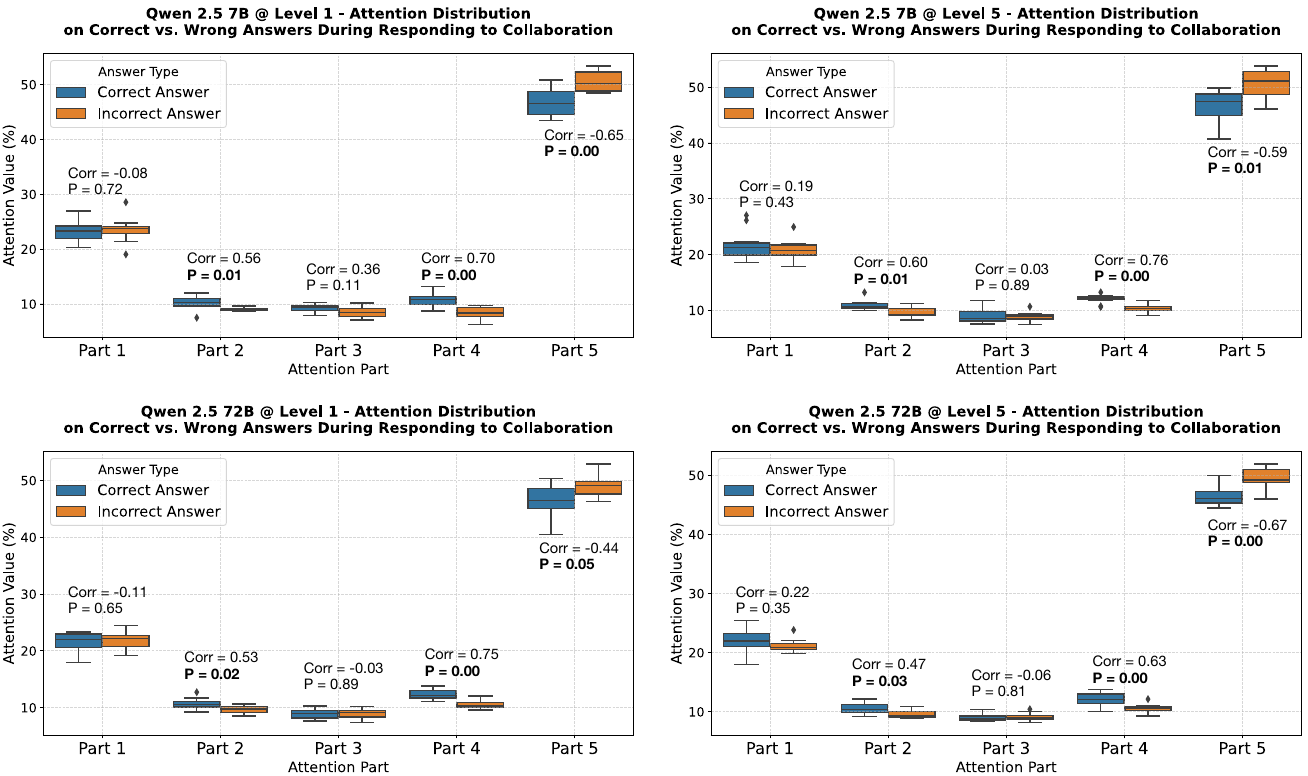}
    }
    \end{subcaptionbox}

    \caption{Attention distributions over different parts of the prompt during successful and failed attempts at initiating and responding to collaboration, evaluated for Qwen-2.5 models (7B and 72B). ``Corr'' denotes the Pearson correlation coefficient between attention patterns and ITES-based success labels, and ``P'' indicates the corresponding p-value.
}
    \label{appendix-corr}
\end{figure*}

Further experimental results are illustrated in Figure~\ref{appendix-corr}, where we report the attention distribution differences across different collaboration scenarios for Qwen-2.5 7B and 72B models at both Level 1 and Level 5. Notably, both Qwen-2.5 7B and 72B exhibit consistent patterns across levels. When initiating collaboration, the attention values on the Collaboration Rule and Recipe parts are significantly correlated with collaboration success or failure. When responding to collaboration, the attention assigned to Collaboration Rule, Environment Observation, and Collaboration Context shows a similar significant correlation.

These findings highlight the critical role of attention mechanisms in LLM-driven collaboration. In particular, the extent to which models attend to collaboration-relevant information is significantly associated with the effectiveness of their collaborative behavior. This relationship holds across different model sizes and task difficulty levels, suggesting a generalizable pattern.

\subsection{Failure Analysis}

\label{ap-fa}

\begin{figure*}[t]
  \includegraphics[width=1\linewidth]{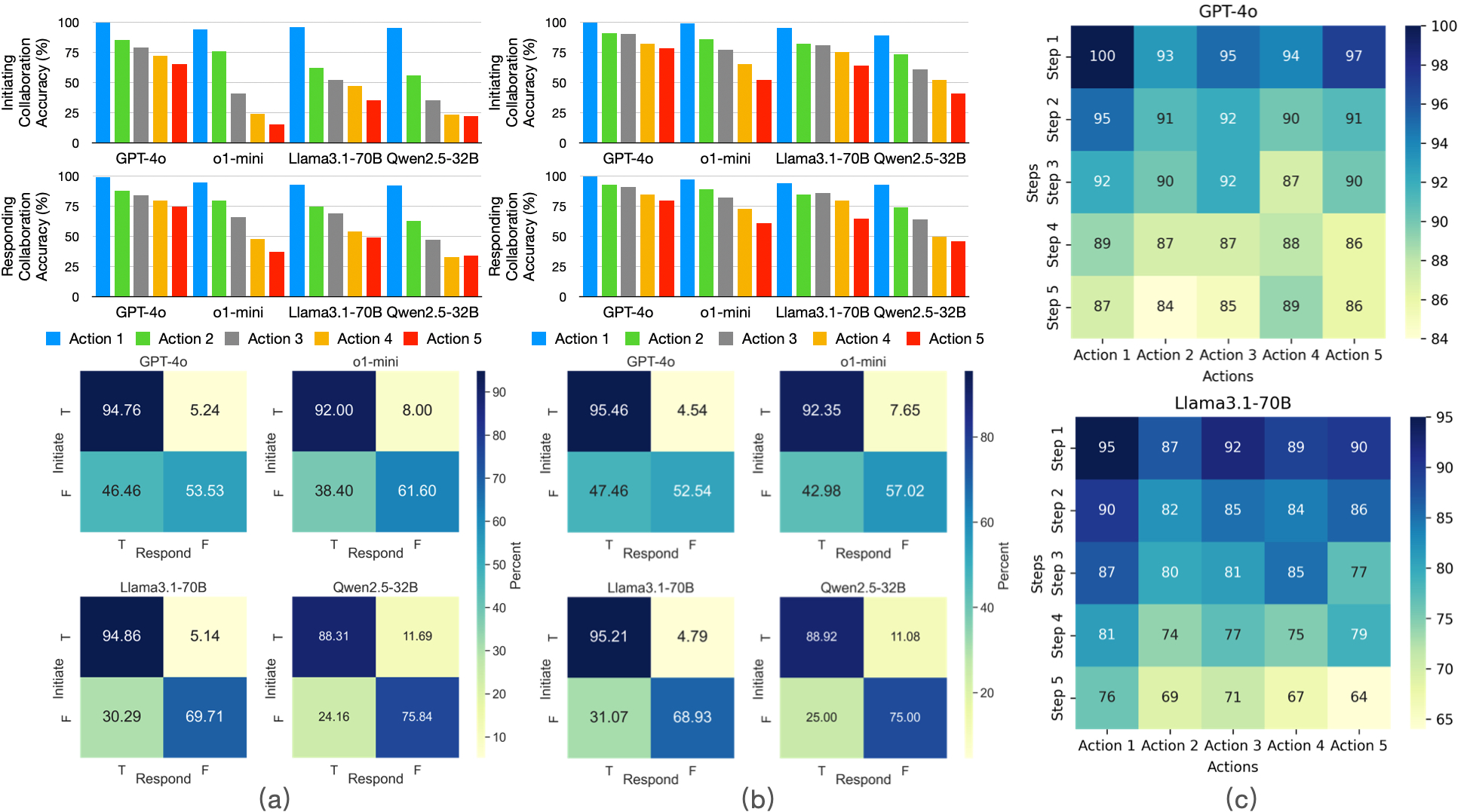}
  \caption {Figure (a) illustrates the dynamic changes in the capabilities of four LLMs in initiating collaboration and responding to collaboration under the original task flow, with the confusion matrix depicting the relationship between the two capabilities. Figure (b) shows the dynamic changes in collaboration capabilities after excluding the impact of task decomposition ability on the task flow. Figure (c) highlights the sensitivity of collaboration capabilities to position, comparing GPT-4o and Llama-3.1-70B after adjusting the position of the task workflow.
}
  \label{fig:fa}
\end{figure*}

\subsubsection{Failure Modes in Collaboration Capabilities Degradation}
\label{ap-new-expa}
To investigate the temporal dynamics and degradation patterns in collaboration capabilities, we designed an experiment focusing on both the initiation and response phases of collaborative actions. Tasks were selected from Level 3, each involving five sequential collaborative actions: ``pickup,'' ``put\_obj\_in\_utensil,'' ``cut/stir,'' ``pickup,'' and ``place\_obj\_on\_counter.'' These actions require implicit collaboration and are not parameterized in advance, as their specifics vary across task instances.

We selected 4 representative LLMs and evaluated them on these five collaborative actions by constructing prompts from environmental states and memory fragments sampled from the agents’ interaction trajectories. For each collaborative action, five representative scenarios were extracted, and each model was tested 20 times per scenario using prompts identical to those in Section~\ref{result}.

\begin{figure*}[t]
  \includegraphics[width=1\linewidth]{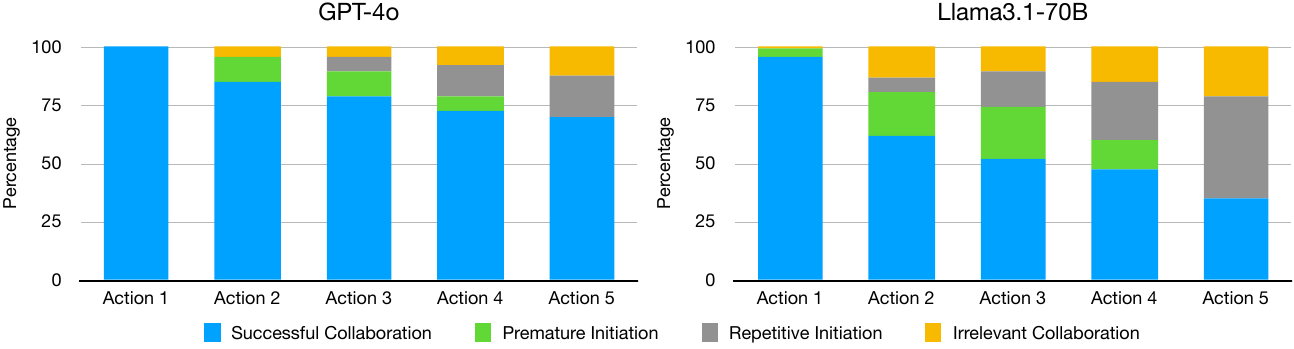}
  \caption {The error condition of GPT-4o and Llama-3.1-70B initiating collaboration.}
  \label{appendix-expa}
\end{figure*}

Collaborative success was measured using the ITES function, where an ITES score greater than 0 was considered a successful action. Failures were manually categorized for initiating agents into three distinct error types, and their distribution is shown in Figure~\ref{appendix-expa}.

\begin{itemize}
    \item[$\bullet$] Premature initiation, where the model attempts a collaborative action before the appropriate task stage;
    \item[$\bullet$] Repetitive initiation, where the model redundantly issues a collaborative action that should have already occurred;
    \item[$\bullet$] Irrelevant collaboration, where the action does not align with any expected collaboration behavior for the task.
\end{itemize}

As illustrated in Figure~\ref{fig:fa}(a), all models performed reliably on the first collaborative action. However, performance declined in subsequent steps. Notably, GPT-4o and Llama-3.1-70B exhibited increasing frequencies of premature and repetitive initiation errors, particularly in later actions. This degradation is more prominent in the smaller Llama-3.1-70B model. This trend is consistent with findings from \cite{li2024measuring}.

Additionally, a confusion matrix analysis revealed a strong dependency between initiation and response behaviors: inaccurate initiation often leads to failed responses. This supports the conclusion that initiation capability is the primary bottleneck in sustaining effective collaboration across temporally extended tasks. The underlying issue appears to be a misalignment between the environmental state and the task’s process-specific progression, which LLM agents may struggle to track consistently without explicit temporal grounding.

\subsubsection{Impact of Task Decomposition Ability}
\label{ap-new-expb}

To further investigate the phenomenon of collaboration capabilities degradation observed in sequential, process-specific tasks, we designed an experiment corresponding to Figure~\ref{fig:fa}(b). This experiment aims to isolate the influence of planning and test whether the decline in collaboration effectiveness is purely due to poor step tracking or is also affected by insufficiently grounded task representations during long-horizon planning.

Building upon the same task setting as Section~\ref{ap-new-expa}, which involved five collaborative actions within Step 1 of a Level 3 task, we redesigned the task recipes to incorporate explicit step-to-action mappings. This allows each step in the recipe to correspond directly to a single collaborative action, thus removing ambiguity in planning. An example of such a reformulated recipe for the "Baked Bell Pepper" task is shown in Listing~\ref{lst-expb}.

Compared to the original recipe structure used in Section~\ref{ap-new-expa}, this revised version decomposes Step 1 into five clear sub-steps, each requiring a distinct and ordered collaborative action. This explicit alignment between steps and actions was designed to eliminate ambiguity in high-level plan formulation, allowing the model to focus on action execution rather than inferring latent step boundaries.

\begin{lstlisting}[caption={Step-to-action mapping recipe of "Baked Bell Pepper"},label={lst-expb}]
NAME:
Baked Bell Pepper

INGREDIENTS:
bell pepper(1)

COOKING STEPs:
1. Pick up a bell pepper.    
2. Place bell pepper on chopping board.   
3. Cut a bell pepper into slices.    
4. Pick up bell pepper slices.    
5. Place the bell pepper slices on counter.
6. Place the bell pepper slices in the oven and bake for 3 timesteps.  
7. Transfer the baked bell pepper slices to a pot and cook for 3 timesteps.
8. Fill a dish with the soup from the pot and serve.
\end{lstlisting}

However, as shown in Figure~\ref{fig:fa}(b), despite this controlled setup, our results show that collaboration capability still declines as the task progresses through the action sequence. This suggests that planning ambiguity is not the sole cause of degradation. Rather, the observed performance drop, particularly in later steps, is likely due to pretraining biases that favor early-stage completions and the model's limited ability to maintain coherent context representations across longer action chains.

Together with the findings of Section~\ref{ap-new-expa}, this experiment reinforces our hypothesis that sequential dependencies and temporal tracking remain key challenges for LLM agents in multi-step collaborative settings, even under explicit instruction-following scenarios.

\subsubsection{Sequence Dependence in Collaboration Performance}

\label{ap-new-expc}

To examine the extent to which collaboration performance is influenced by step position rather than content or complexity, we conducted an experiment corresponding to Figure~\ref{fig:fa}(c). This experiment builds directly upon the structure of Section~\ref{ap-new-expb}, which provided explicit step-to-action mappings, and focuses on determining whether poor performance in later steps is attributable to their position in the sequence rather than inherent task complexity.

We reordered the steps of the ``Baked Bell Pepper'' recipe such that each collaborative action previously occurring later in the sequence was moved to Step 1. The goal was to evaluate whether this positional shift would lead to improved performance for actions that previously suffered from degradation. Listing~\ref{lst-expc} presents an example where the action originally in Step 2 (i.e., place bell pepper on chopping board) is now assigned to Step 1. For clarity, the square brackets annotate the original step numbers and were not visible to models during the experiment.

The results demonstrated a significant increase in collaboration performance when previously underperforming actions were moved to earlier steps. Actions that had shown degradation in their original later positions now performed comparably to the original Step 1, and the overall pattern of performance decline across the sequence largely disappeared.

\begin{figure}[H]
\begin{lstlisting}[caption={Rearranged recipe of "Baked Pumpkin Soup"},label={lst-expc}]
NAME:
Baked Pumpkin Soup

INGREDIENTS:
bell pepper(1)

COOKING STEPs: 
[Previously for step 2]
1. Place bell pepper on chopping board.
[Previously for step 3]
2. Cut a bell pepper into slices.
[Previously for step 4]
3. Pick up bell pepper slices.
[Previously for step 5]
4. Place the bell pepper slices on counter.
[Previously for step 1]
5. Pick up a bell pepper.
[The following are not the steps corresponding to collaborative action]
6. Place the bell pepper slices in the oven and bake for 3 timesteps.
7. Transfer the baked bell pepper slices to a pot and cook for 3 timesteps.
8. Fill a dish with the soup from the pot and serve. 
\end{lstlisting}
\end{figure}

These findings suggest that the observed degradation in collaborative capabilities is not solely due to action difficulty or planning ambiguity but is strongly influenced by positional effects. This positional dependence may stem from two key factors: (1) Pretraining biases in LLMs that favor earlier sequence completions (e.g., next-token prediction dominance at sequence heads), and (2) Limited ability to maintain coherent task context across extended action chains, especially when no explicit memory or reasoning loop is enforced. By isolating position as a variable, it is demonstrated that early-sequence placement alone can substantially boost performance in collaborative tasks, highlighting a structural limitation in current LLM planning and grounding mechanisms when applied to long-horizon collaboration.

\subsection{Case Study}

We present case studies of agent collaboration processes, using the DeepSeek-V3 model to illustrate four scenarios: successful initiating and responding, successful initiating but failed responding, failed initiating but successful responding, and failed initiating and responding. For each case, we provide the agent’s environmental state inputs, along with the output of the agent, including the agent's analysis, dialogue, and collaborative actions.

\paragraph{Case 1: Successful Initiating and Responding}
Listing \ref{lst-1-1}, \ref{lst-1-2}, and \ref{lst-1-3} present the input from Agent Bob and the output from Agent Alice in Case 1. To prepare the order, Bob requires Alice's assistance in obtaining ingredients for further processing. As indicated by Bob's analysis field, it correctly understood the first step of the recipe, which is to retrieve a bell pepper from the ingredient dispenser. Furthermore, Bob recognized that it could not access the dispenser and thus requested Alice to retrieve the bell pepper on its behalf. Ultimately, Bob correctly initiated the collaborative action by requesting Alice to take the bell pepper from the ingredient dispenser and place it on the counter. Therefore, Bob's initiation was accurate. Similarly, Alice's plan field correctly generated the two collaborative actions initiated by Bob, which will be executed subsequently. As a result, Alice appropriately responded to the collaboration.

\begin{figure}[H]
\begin{lstlisting}[caption={Case 1 Agent Bob's input},label={lst-1-1}]
Successful Action History: []
Bob space:pot0  oven0  counter 
Alice space:chopping_board0  blender0  dish_dispenser  ingredient_dispenser
Order:baked_bell_pepper
Scene 0: <Bob> holds nothing. The planned sequence of actions (yet to be performed) for Bob is [] <Alice> holds nothing. The planned sequence of actions (yet to be performed) for Alice is [] Kitchen states: <pot0> is empty; <chopping_board0> is empty; <oven0> is empty; <blender0> is empty; 3 counters can be visited by <Bob>. Their states are as follows: counters have nothing
\end{lstlisting}
\end{figure}

\begin{figure}[H]

\begin{lstlisting}[caption={Case 1 Agent Bob's output},label={lst-1-2}]
Bob analysis: The order is for a baked bell pepper. According to the recipe, the first step is to pick up a bell pepper. Since only the Alice can pick up ingredients from the ingredient dispenser, I need to request the Alice to perform this action. Once the bell pepper is on the counter, I can proceed to place it in the oven and bake it.
Bob plan: request('pickup(bell_pepper, ingredient_dispenser)'); request('place_obj_on_counter()')
Bob say: Alice, please pick up a bell pepper from the ingredient dispenser and place it on the counter. [END]
\end{lstlisting}
\end{figure}

\begin{figure}[H]

\begin{lstlisting}[caption={Case 1 Agent Alice's output},label={lst-1-3}]
Alice analysis: The Bob has instructed me to pick up a bell pepper from the ingredient dispenser and place it on the counter. Since I am the Alice and only I can pick up ingredients from the ingredient dispenser, I will follow the Bob's instructions. The ingredient dispenser is in my space, so I can perform this action directly. After picking up the bell pepper, I will place it on the counter so the Bob can access it.
Alice plan: pickup(bell_pepper, ingredient_dispenser); place_obj_on_counter()
Alice say: [NOTHING]
\end{lstlisting}
\end{figure}

\paragraph{Case 2: Failed Initiating but Successful Responding}

Listing \ref{lst-2-1}, \ref{lst-2-2}, and \ref{lst-2-3} present the input from Agent Bob and the output from Agent Alice in Case 2. In this scenario, Bob was supposed to request Alice to retrieve the pumpkin from the ingredient dispenser and place it on the chopping board (chopping\_board0) for cutting. However, before requesting Alice to ``cut'', Bob neglected to generate the action ``put\_obj\_in\_utensil(chopping\_board0)'' to place the ingredient on the chopping board. Since the ingredient must be placed on the chopping board before it can be cut, Bob's collaborative request was incorrect. Although Alice recognized Bob’s intention, it identified the missing action ``put\_obj\_in\_utensil(chopping\_board0)'' during its planning phase and included it. This correction enabled the successful completion of the task of cutting the pumpkin. Therefore, Alice's response was successful, as it corrected Bob's mistake.

\begin{figure}[H]

\begin{lstlisting}[caption={Case 2 Agent Bob's input},label={lst-2-1}]
Successful Action History: []
Bob space:pot0  oven0  counter 
Alice space:chopping_board0  blender0  dish_dispenser  ingredient_dispenser
Order:sliced_pumpkin_and_chickpea_stew
Scene 11: <Bob> holds nothing. The planned sequence of actions (yet to be performed) for Bob is [] <Alice> holds nothing. The planned sequence of actions (yet to be performed) for Alice is [pickup(pumpkin, ingredient_dispenser),place_on_counter(),pickup(chickpea, ingredient_dispenser)] Kitchen states: <pot0> is empty; <chopping_board0> is empty; <oven0> is empty; <blender0> is empty; 3 counters can be visited by <Bob>. Their states are as follows: counters have nothing.

\end{lstlisting}
\end{figure}

\begin{figure}[H]

\begin{lstlisting}[caption={Case 2 Agent Bob's output},label={lst-2-2}]
Bob analysis: The recipe for Sliced Pumpkin and Chickpea Stew requires us to first cut the pumpkin into slices, then place the pumpkin slices and chickpea in a pot, cook them, and finally fill a dish with the stew and serve. The Alice is currently planning to pick up the pumpkin and chickpea from the ingredient dispenser. Once the Alice has the ingredients, they will need to place them on the counter for me to handle. I will need to pick up the pumpkin from the counter and cut it into slices using the chopping board, but since the chopping board is in the Alice's space, I will need to request them to perform the cutting action.  
Bob plan: request('cut(chopping_board0)'); wait(1)
Bob say: Alice, please cut the pumpkin on the chopping board after picking it up. [END]
\end{lstlisting}
\end{figure}

\begin{figure}[H]
\begin{lstlisting}[caption={Case 2 Agent Alice's output},label={lst-2-3}]
Alice analysis: Based on the Bob's instruction in the conversation history, I need to pick up the pumpkin from the ingredient dispenser and then cut it on the chopping board. The Alice space includes a chopping_board0, which is available for use. Since the Bob has not provided further instructions, I will focus on these two tasks. The planned sequence of actions for the Alice already includes picking up the pumpkin and chickpea, but the Bob specifically mentioned cutting the pumpkin first. Therefore, I will prioritize picking up the pumpkin and cutting it.
Alice plan: pickup(pumpkin, ingredient_dispenser), put_obj_in_utensil(chopping_board0),cut(chopping_board0)
Alice say: [NOTHING]
\end{lstlisting}
\end{figure}

\paragraph{Case 3: Successful Initiating but Failed Responding}

Listing \ref{lst-3-1}, \ref{lst-3-2}, and \ref{lst-3-3} present the input from Agent Bob and the output from Agent Alice in Case 3. In this scenario, Bob correctly requested Alice to retrieve the cauliflower from the ingredient dispenser and place it on the chopping board (chopping\_board0) for cutting, and the collaborative actions generated were also correct. Therefore, the initiation of collaboration was accurate. However, Alice's first action, ``pickup(cauliflower, dispenser)'', contained an incorrect parameter, as ``dispenser'' was not a valid option. This error in the first action prevented the execution of subsequent actions, resulting in an incorrect response to the collaboration.

\begin{figure}[H]
\begin{lstlisting}[caption={Case 3 Agent Bob's input},label={lst-3-1}]
Successful Action History: []
Bob space:pot0  oven0  counter 
Alice space:chopping_board0  blender0  dish_dispenser  ingredient_dispenser
Order:mashed_cauliflower_and_lentil_patty
Scene 0: <Bob> holds nothing. The planned sequence of actions (yet to be performed) for Bob is [] <Alice> holds nothing. The planned sequence of actions (yet to be performed) for Alice is [] Kitchen states: <pot0> is empty; <chopping_board0> is empty; <oven0> is empty; <blender0> is empty; 3 counters can be visited by <Bob>. Their states are as follows: counters have nothing.
\end{lstlisting}
\end{figure}

\begin{figure}[H]
\begin{lstlisting}[caption={Case 3 Agent Bob's output},label={lst-3-2}]
Bob analysis: According to the recipe, we need to start by cutting the cauliflower into slices. Since I don't have access to the chopping board or the ingredient dispenser, I need to ask the Alice to pick up the cauliflower and cut it. The Alice can perform these actions in their space. Once the cauliflower is sliced, it needs to be placed in a pot to cook, which is in my space. I will request the Alice to pick up the cauliflower and cut it, then place it on the counter so I can transfer it to the pot.  
Bob plan: request('pickup(cauliflower, ingredient_dispenser)'); request('put_obj_in_utensil(chopping_board0)');request('cut(chopping_board0)'); request('place_obj_on_counter()')  
Bob say: Please pick up the cauliflower from the ingredient dispenser, cut it on the chopping board, and place it on the counter for me to handle next. [END]
\end{lstlisting}
\end{figure}

\begin{figure}[H]
\begin{lstlisting}[caption={Case 3 Agent Alice's output},label={lst-3-3}]
Alice analysis: The Bob has instructed me to pick up the cauliflower from the ingredient dispenser, cut it on the chopping board, and place it on the counter. I need to follow this sequence since I do not have the recipe and must rely on the Bob's guidance. Currently, I am not holding anything, and all utensils in my space are empty. I will start by picking up the cauliflower from the ingredient dispenser, then proceed to cut it on the chopping board, and finally place it on the counter for the Bob to handle next.
Alice plan: pickup(cauliflower, dispenser);put_obj_in_utensil(chopping_board0); cut(chopping_board0); place_obj_on_counter()
Alice say: [NOTHING]
\end{lstlisting}
\end{figure}

\paragraph{Case 4: Failed Initiating and Responding}

Listing \ref{lst-4-1}, \ref{lst-4-2}, and \ref{lst-4-3} present the input from Agent Bob and the output from Agent Alice in Case 4. In this scenario, Bob requested Alice to cut the eggplant on ``chopping\_board0'' and place it on the counter. However, the collaborative actions generated were incomplete, as they omitted the actions ``put\_obj\_in\_utensil(chopping\_board0)'' and ``place\_obj\_on\_counter()''. As a result, these actions could not be executed, making them incorrect. Although Alice understood Bob's intention, it failed to notice the missing actions in Bob's request and directly replicated his generated actions. Consequently, the response to the collaboration was also incorrect, leading to a failure in the collaboration.

\begin{figure}[H]
\begin{lstlisting}[caption={Case 4 Agent Bob's input},label={lst-4-1}]
Successful Action History: []
Bob space:pot0  oven0  counter 
Alice space:chopping_board0  blender0  dish_dispenser  ingredient_dispenser
Order:sliced_eggplant_and_chickpea_stew
Scene 4: <Bob> holds nothing. The planned sequence of actions (yet to be performed) for Bob is [] <Alice> holds one eggplant. The planned sequence of actions (yet to be performed) for Alice is [request('place_obj_on_counter()')] Kitchen states: <pot0> is empty; <chopping_board0> is empty; <oven0> is empty; <blender0> is empty; 3 counters can be visited by <Bob>. Their states are as follows: counters have nothing.
\end{lstlisting}
\end{figure}

\begin{figure}[H]
\begin{lstlisting}[caption={Case 4 Agent Bob's output},label={lst-4-2}]
Bob analysis: The recipe requires preparing a Sliced Eggplant and Chickpea Stew. The Alice currently holds an eggplant and plans to place it on the counter. I need to instruct the Alice to cut the eggplant into slices and then place them in the pot, along with the chickpea. Since the Alice is in charge of handling the ingredients, I will focus on managing the cooking process in the pot and ensuring the dish is served correctly.
Bob plan: request('cut(chopping_board0)'); request('place_obj_on_counter()')
Bob say: Please cut the eggplant into slices using the chopping board and then place the slices on the counter. [END]
\end{lstlisting}
\end{figure}

\begin{figure}[H]
\begin{lstlisting}[caption={Case 4 Agent Alice's output},label={lst-4-3}]
Alice analysis: I am currently holding an eggplant, and the Bob has instructed me to cut it into slices using the chopping board and then place the slices on the counter. The chopping board is in my space, so I can perform the cutting action. After cutting, I will place the slices on the counter for the Bob to use in the next steps.
Alice plan: cut(chopping_board0);place_obj_on_counter()
Alice say: [NOTHING]
\end{lstlisting}
\end{figure}

\end{document}